\documentclass{article}

\usepackage{arxiv}

\usepackage[utf8]{inputenc} % allow utf-8 input
\usepackage[T1]{fontenc}    % use 8-bit T1 fonts

\usepackage{url}            % simple URL typesetting
\usepackage{booktabs}       % professional-quality tables
\usepackage{amsfonts}       % blackboard math symbols
\usepackage{nicefrac}       % compact symbols for 1/2, etc.
\usepackage{microtype}      % microtypography
\usepackage{graphicx}
\graphicspath{ {./images/} }

\usepackage{amssymb}
\usepackage{amsmath}
\usepackage{amsthm}

\usepackage{natbib}
\usepackage{color}
\usepackage{array}
\newcolumntype{M}[1]{>{\centering\arraybackslash}m{#1}}
\usepackage{pythonhighlight}
\usepackage{tcolorbox}
\usepackage{listings}
\usepackage{hyperref}

\usepackage{mathtools}
\usepackage{subcaption}
\usepackage{multicol}

\title{\textit{VICCA}: Visual Interpretation and Comprehension of Chest X-ray Anomalies in Generated Report Without Human Feedback}

\author{{\hspace{1mm}Sayeh GHOLIPOUR PICHA}\\
	Univ. Grenoble Alpes,\\
        CNRS, Grenoble INP, GIPSA-lab,\\
        38000 Grenoble, France\\
	\texttt{sayeh.gholipour-picha@grenoble-inp.fr} \\
	\And
	{\hspace{1mm}Dawood AL CHANTI} \\
	Univ. Grenoble Alpes,\\
        CNRS, Grenoble INP, GIPSA-lab,\\
        38000 Grenoble, France\\
	\texttt{dawood.al-chanti@grenoble-inp.fr} \\
        \And
        {\hspace{1mm}Alice CAPLIER} \\
	Univ. Grenoble Alpes,\\
        CNRS, Grenoble INP, GIPSA-lab,\\
        38000 Grenoble, France\\
	\texttt{alice.caplier@grenoble-inp.fr} \\
}

\begin{document}

\maketitle
%% Abstract
\begin{abstract}
As artificial intelligence (AI) becomes increasingly central to healthcare, the demand for explainable and trustworthy models is paramount. Current report generation systems for chest X-rays (CXR) often lack mechanisms for validating outputs without expert oversight, raising concerns about reliability and interpretability. To address these challenges, we propose a novel multimodal framework designed to enhance the semantic alignment between text and image context and the localization accuracy of pathologies within images and reports for AI-generated medical reports.
Our framework integrates two key modules: a Phrase Grounding Model, which identifies and localizes pathologies in CXR images based on textual prompts, and a Text-to-Image Diffusion Module, which generates synthetic CXR images from prompts while preserving anatomical fidelity. By comparing features between the original and generated images, we introduce a dual-scoring system: one score quantifies localization accuracy, while the other evaluates semantic consistency between text and image features.
Our approach significantly outperforms existing methods in pathology localization, achieving an 8\% improvement in Intersection over Union score. It also surpasses state-of-the-art methods in CXR text-to-image generation, with a 1\% gain in similarity metrics. Additionally, the integration of phrase grounding with diffusion models, coupled with the dual-scoring evaluation system, provides a robust mechanism for validating report quality, paving the way for more reliable and transparent AI in medical imaging.
\end{abstract}

% keywords can be removed

\section{Introduction}
\label{sec:intro}
\begin{figure}
    \centering
    \includegraphics[width=\linewidth]{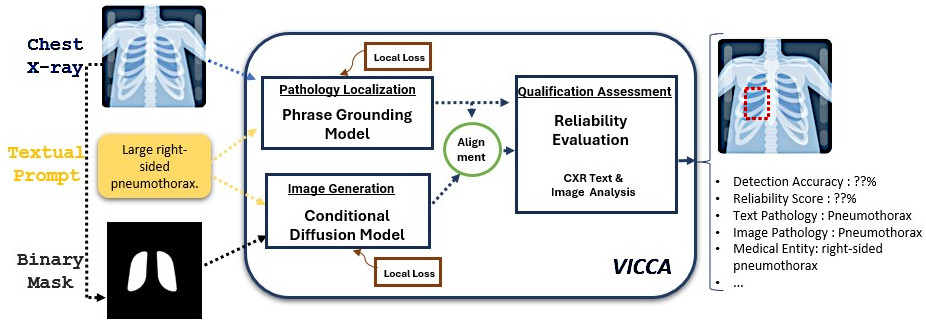}
    \caption{Overview of VICCA. Given a chest X-ray image and a text query, the model performs visual localization of the text input and generates a comprehensive set of information. This includes the localization accuracy, the alignment accuracy between the visual output and the text query, the presence of the specified pathology in both the text query and the chest X-ray image, the identification of the main entity within the text prompt, and a summarization of the text input into its key medical entities.}
    \label{fig:vicca}
\end{figure}

Recent advancements in using multimodal data by integrating textual data with images in the medical domain have revolutionized diagnostic workflows in AI-driven healthcare \cite{Yildirim2024MultimodalHA1, chen2024artificial_Multimodal2}, mirroring trends in natural image analysis.
Among these advancements, large language models (LLMs) and Visual Language Models (VLMs) have shown considerable growth in processing and generating text from text and image, excelling in applications such as conversational AI and image captioning. 
Despite the popularity of LLMs and VLMs in natural image processing, their application in medical imaging remains in its early stages, with only a handful of studies exploring this synergy in the medical field \cite{muller2024chex, bannur2024maira-}.
For instance, using VLMs in medical image captioning using text and image data has shown great potential for enhancing diagnostic accuracy \cite{bannur2024maira-}, enabling richer contextual understanding \cite{muller2024chex}, and supporting more precise decision-making. These models offer the potential to generate detailed and contextually relevant medical reports from images.
However, unlike image captioning in natural data, medical reports require an in-depth understanding of domain-specific terminology, the ability to interpret subtle pathological features, and an alignment with structured medical reporting standards \cite{image_captioning1, image_captioning2, boecking2022making, gloria} to ensure the generated information is clinically relevant and contextually accurate. 
Nonetheless, using textual information alongside medical images introduces additional complexity, potentially increasing the challenges of interpretability and trust in automated systems as errors or ambiguities in generated reports could lead to misdiagnosis or mistrust among practitioners \cite{Yildirim2024MultimodalHA1}.
This complexity underscores the need for innovative frameworks to ensure reliability and facilitate validation. 

Addressing the lack of quantifiable validation metrics in AI-generated medical reports, such as a metric that ensures alignment and consistency between textual and visual data, would provide a more robust foundation for clinicians to evaluate and rely on these outputs. Even though integrating multimodal data introduces additional complexity, it also enables the development of a double-mapping approach that links information across modalities, which would facilitate a deeper understanding of the relationship between image features and corresponding textual descriptions.

While our work focuses on chest X-rays (CXR) due to the availability of datasets and the critical need for advancements in radiology, LLMs and VLMs have also been applied in other medical domains, such as MRI for brain tumor analysis and ultrasound for fetal health assessment. These applications highlight the growing importance of multimodal AI in healthcare, showing its potential to integrate various data sources for improved diagnostic insights. The framework we propose offers a generalizable solution that could be extended to other modalities, further advancing AI's role in healthcare.

In the context of Chest X-ray analysis, few studies \cite{muller2024chex, tanida2023interactive, bannur2024maira-} have focused on improving automated report generation by detecting pathologies in CXR images and producing structured, clinically relevant reports based on identified regions of interest (ROI). However, these AI-generated reports are often not self-explanatory and do not include standardized metrics to evaluate their trustworthiness. Consequently, expert validation remains imperative, and the trustworthiness of these reports continues to be a significant concern. 
This issue is further complicated by the variability in current methods for CXR report generation, particularly in how differently they contextualize and structure content to address clinical terms.
Differences in the choice of training datasets and model architectures can lead to significant variations in generated outputs \cite{cxr-repair, tanida2023interactive}. For instance, one model might generate a report indicating 'no findings' for a given image, while another may identify and describe a pathology.

To address these limitations in AI-generated medical reports, our research introduces a novel and comprehensive multimodal framework that enhances the visual interpretability, reliability, and consistency of generated reports. Unlike previous approaches \cite{muller2024chex, bannur2024maira-}, our framework not only integrates multimodal data but also introduces a unique evaluation metric designed to quantify the quality and reliability of generated reports, as well as their alignment with the corresponding medical images. Our approach leverages report content to localize pathologies within imaging data and validates these localizations through a complementary AI module, which generates synthetic images from text data. By comparing the consistency of ROIs from the localization across modalities, the framework produces quantitative confidence scores, reinforcing the reliability of the outputs.

\textit{Our multi-modalities model operates as follows}: 
The phrase grounding model processes a CXR image along with a corresponding text prompt describing the pathologies. It outputs bounding boxes that localize the identified pathologies within the image. Subsequently, a conditional diffusion model takes the same text prompt and a binary anatomical mask derived from the original CXR image as input to generate a new synthetic CXR image. This generated image retains the original anatomical structure while incorporating the pathology described in the prompt.

The quality of the text prompt significantly affects the accuracy of the generation. When the report provides rich and detailed attributes, such as precise location or clarification (e.g., distinguishing opacity from pulmonary edema or fibrosis), the generation model produces more realistic and accurate outputs. However, for some conditions like cardiomegaly (enlarged heart) or cardiomediastinal contour (heart contours), the model often learns the spatial context during training and can reliably associate these findings with the heart region, as shown in the second row of Table \ref{tab:sample_CXRs}. Similarly, pathologies like communicating pneumothorax are consistently linked to chest wall regions.

In contrast, for findings such as pneumonia or pulmonary edema, location within the lung fields becomes critical. The third row of Table \ref{tab:sample_CXRs} illustrates how these pathologies can appear visually similar at the image level. In such cases, the clarity and specificity of the report directly influence the fidelity of the generated image and, by extension, the reliability of subsequent feature analyses. Therefore, evaluating how well the text aligns with the original image provides a meaningful way to assess the quality and trustworthiness of the report.

\begin{table}[t!]
    \centering
    \caption{Sample chest X-ray (CXR) images. The top row shows a normal CXR, while the remaining rows display CXRs with various pathologies, including right-sided cardiomegaly, large pneumothorax, viral pneumonia, and pulmonary edema.}
    \begin{tabular}{|c|c|}
    \hline
    \multicolumn{2}{|c|}{Normal Chest X-ray Image}\\
    \hline
    \multicolumn{2}{|c|}{\includegraphics[width=0.2\textwidth]{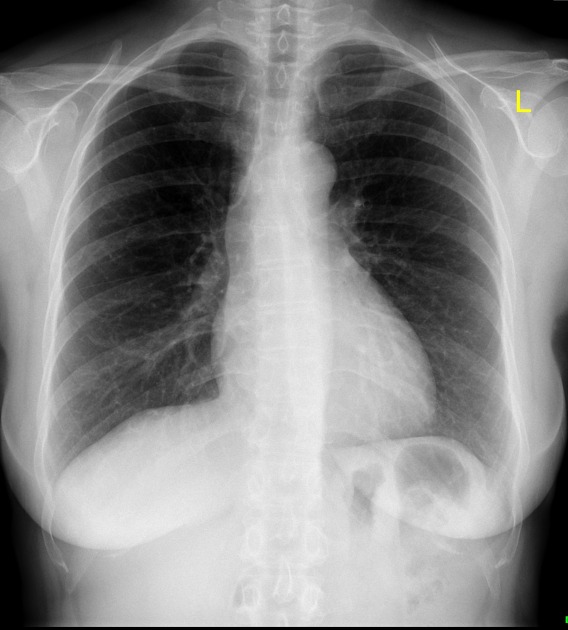}}\\
    \hline
    Right-sided Cardiomegaly & Large Pneumothorax\\
    \hline
    \includegraphics[width=0.2\textwidth]{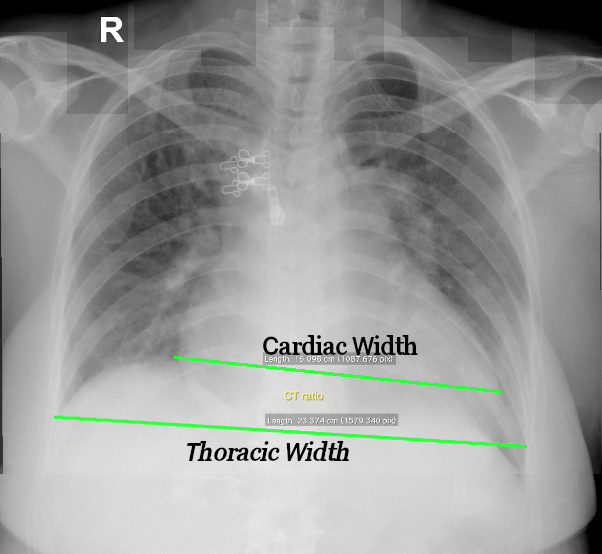} & \includegraphics[width=0.2\textwidth]{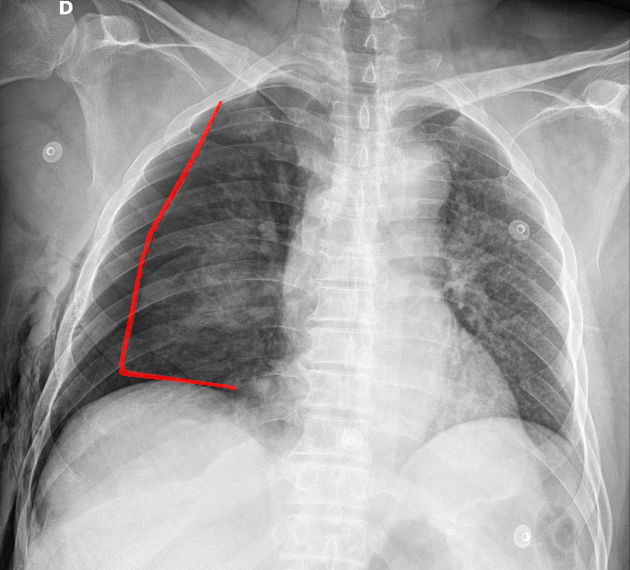}\\
    \hline
    Virus Pneumonia & Pulmonary Edema\\
    \hline
    \includegraphics[width=0.2\textwidth]{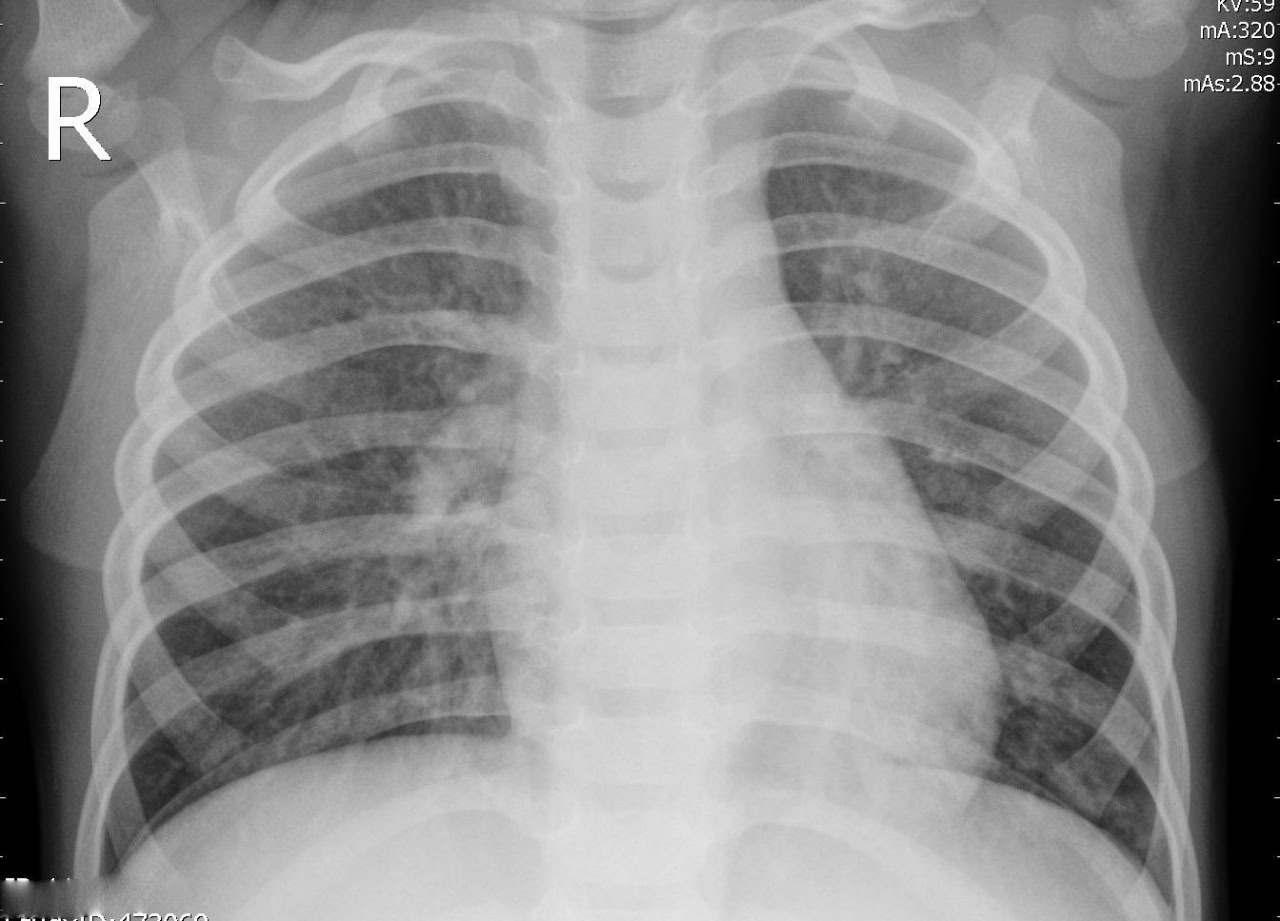} &
    \includegraphics[width=0.2\textwidth]{table_1_5_pulmonary_edema.jpg}\\
    \hline
    \end{tabular}
    \label{tab:sample_CXRs}
\end{table}

To evaluate the consistency and accuracy of the generated outputs, we register the previously detected bounding boxes onto the synthetic image and assess the ROIs' compatibility using feature-based similarity metrics. This process results in two distinct scores:
\begin{enumerate}
    \item \textbf{Detection Accuracy:} Quantifies the likelihood that the localized region corresponds to the textual content, indicating how confidently the \textbf{phrase grounding model} associates the detected pathology with the context of the report.
    % Quantifies the probability of the alignment between the localized regions and the report content, ensuring accurate text-to-image correspondence, and providing a measure of the visualization validity for text-to-image localization.
    \item \textbf{Reliability Score:} A new interpretability score verifies whether the localized features align with the semantic features of the report. This is achieved through text-to-image generation techniques that produce new CXR images while preserving anatomical structure. Assessing the feature consistency of the ROIs between generated and original images which contains the most important information regarding the anomaly/pathology.
\end{enumerate}
The rationale behind this approach lies in detecting potential errors in the generated reports. If the report inaccurately describes features not present in the original CXR image, these discrepancies will manifest in the synthetic image generated from the report's content. Consequently, the compatibility between the original and synthetic images in their respective ROIs will be low, signaling potential inaccuracies in the report's generated features. This dual-scoring mechanism ensures a robust evaluation of both localization precision and semantic alignment, enhancing the trustworthiness of AI-generated reports.

In our approach, to address the lack of external radiologist inputs, we design an automated validation system for our pipeline that does not rely on expert feedback, addressing challenges such as limited accessibility, high costs, and the occasional subjectivity of expert opinions. Instead, we leverage annotated datasets and established evaluation metrics to ensure an objective and reproducible validation process. This system is adaptable across diverse databases, enhancing the generalizability and robustness of our model evaluation.
Furthermore, our framework validates each model in the pipeline individually, focusing on their medical understanding, covering both anatomy and pathology and their overall performance.

The output of our pipeline also includes additional spatial information about the image pathology, derived from the TorchXRayVision model \cite{Cohen2022xrvtorchxrayvision}. Furthermore, it extracts text pathology and medical entities from the input text using our previous model \cite{MCSE}, comparing the interpretability of both the visual and textual data.

Our primary contribution is the introduction of a novel cross-validating multimodal pipeline that can be broken down into the following parts, as shown in Figure \ref{fig:vicca}:

\begin{enumerate}
    \item \textbf{Pathology Localization Model:} We fine-tune and modify a phrase grounding model originally trained on natural image datasets to adapt its architecture for medical chest X-ray applications. To enhance its performance, we integrate a BERT-like encoder \cite{BertDevlin2019BERTPO, boecking2022making}, pre-trained on CXR report texts, and further optimize it to complement the architecture effectively.
    
    \textit{Output:}The model produces bounding boxes indicating the locations of the detected findings.
    
    \item \textbf{Image Generation:} A stable diffusion model is developed for text-to-image generation \cite{11DiffusionNEURIPS2021_49ad23d1, zhang2023adding_controlnet}. This model generates CXRs from input text reports while preserving the anatomical structure of the original image. To ensure anatomical integrity, a binary mask segmented from the lung regions of the original CXR guides the generation process, maintaining alignment with the underlying medical context.
    
    \textit{Output:} The synthetic image closely resembles the original CXR while integrating key features described in the input text report.

    \item\textbf{Medical Validation:} A validation method is developed to assess the anatomical structure and pathological accuracy of the generated CXRs. This is achieved either by leveraging existing well-known models \cite{Cohen2022xrvtorchxrayvision} or by training specialized AI models \cite{detr} with high accuracy customized to the task. The approach ensures that the generated images are both realistic and clinically relevant.
    \item \textbf{Qualification Assessment:} To assess the similarity of regional bounding boxes between the original and generated images, we introduce a false report with no resemblance to the original report in thoracic disease and use it to generate CXR images. The goal is to observe whether the distribution of features changes as a result.

\end{enumerate}

By making these contributions, we advance the integration of multimodal models in medical imaging, thereby enhancing both the generation and interpretability of medical reports and images.

\section{Related Work}
\label{sec:related_work}
In this project, we introduce a novel pipeline for interpretability in chest X-ray report generation. For better comparison, we break down the pipeline into key components: i) Medical Phrase Grounding, section \ref{subsec:MPG} and ii) Chest X-ray Text-to-Image Diffusion, section \ref{subsec:CXR_diffusion}, corresponding to methods used in related studies. We will next discuss each module and its limitations within the medical context then compare our results with these individual components and validate the pipeline's final output accordingly. 

\subsection{Medical Phrase Grounding}
\label{subsec:MPG}

Phrase grounding models focus on locating specific findings or phrases described in text within medical images. Works such as \cite{boecking2022making, Chen_miccai2023, vilouras2024zeroshotmedicalphrasegrounding} demonstrate the utility of these approaches for identifying anomalies. For instance, Boecking et al. \cite{boecking2022making} propose a zero-shot method that maps textual findings to ROIs in CXRs. Similarly, Chen et al. \cite{Chen_miccai2023} highlight challenges in locating small or overlapping findings, particularly where visual and textual cues may conflict.
Despite their progress, these models primarily serve as localization tools and often lack mechanisms to validate their outputs without expert feedback. This limits their reliability in practical clinical settings. Additionally, their generalizability is constrained due to limited medical training data compared to natural image datasets. 
Addressing these limitations, we introduce an automated reliability evaluation score that reduces dependence on expert validation by assessing the semantic alignment between the generated report and the image.

Alternatively, Grounded reporting methods \cite{Ichinose_miccai2023, tanida2023interactive, muller2024chex, zou2024medrg} extend phrase grounding by combining localization with automated report generation. These methods aim to improve the accuracy and clinical relevance of generated reports by first detecting regions of interest and then generating textual descriptions. For example, Tanida et al. \cite{tanida2023interactive} demonstrate that grounded approaches outperform direct image-to-text generation in identifying multiple anomalies.
While grounded reporting improves report precision, the generated reports still face challenges in terms of interpretability and validation. Moreover, these methods primarily focus on generating new reports rather than enhancing existing ones.

In contrast to previous approaches, our work leverages medical phrase grounding to localize findings described in a given CXR report, while focusing on enhancing the interpretability and trustworthiness of existing AI-generated reports.
This method facilitates visual validation and bridges the gap between textual and visual modalities. 
Furthermore, we enhance the interpretability by designing a new dual-scoring system that transforms trust from a binary agree/disagree metric into a continuous, quantifiable measure of performance, thereby enabling greater confidence in the model's outputs.

Unlike models trained predominantly on natural image datasets \cite{42liu2023groundingdino}, our architecture is fine-tuned using medical data, ensuring it is specifically adapted to the unique complexities of CXR images, such as subtle anatomical structures and overlapping pathologies. 
To assess model performance without expert feedback, we integrate a validation pipeline that utilizes annotated datasets and standardized metrics. This provides a robust and objective evaluation framework, ensuring both reliability and clinical relevance in the assessment process.

\subsection{Chest X-ray text to image Diffusion}
\label{subsec:CXR_diffusion}

Text-to-image diffusion models represent a novel approach in medical imaging by generating synthetic CXR images from textual reports. These models operate by mapping text descriptions to visual features, effectively synthesizing images that align with the described pathologies and anatomical context. This process can enhance medical image analysis by addressing issues such as data scarcity, enabling augmentation \cite{pinaya2022brainimaginggenerationlatent}, and supporting research validation \cite{Rao2025-vt}. Despite their promise, current methods face challenges in maintaining anatomical consistency, a critical requirement for clinical applicability.

Pioneering works like RoentGen \cite{6chambon2022roentgen} and Cheff \cite{36cheffweber2023cascaded} have demonstrated the potential of generating CXRs directly from free-form text reports. While these methods can produce plausible images, they often lack control over critical spatial features, resulting in inconsistencies in the anatomical structure. For example, a generated image describing ``left lung consolidation" might misplace the pathology or alter unrelated anatomical regions, thereby reducing clinical reliability \cite{5chambon2022adapting, 32Rombach2021HighResolutionIS}.

Class-conditional X-ray generation approaches, such as \cite{27packhauser2022generation}, address privacy concerns with advanced sampling strategies but primarily focus on dataset augmentation rather than preserving anatomical detail. While useful for specific tasks, such methods do not fully meet the needs of structured, pathology-guided image generation, particularly when spatial accuracy is critical. In our earlier work \cite{my_master_thesis}, we highlighted the importance of using synthetic data to balance datasets effectively, underscoring that, for our research, preserving anatomical structure in chest X-rays is essential, particularly when incorporating spatial information like pathology. Recent studies have further explored the use of generative adversarial networks (GANs) to enhance the robustness of medical predictions under imbalanced data conditions, reinforcing the potential role of synthetic data in clinical decision-making pipelines \cite{suggested_cite_1}.

A more recent study, XReal \cite{hashmi2024xreal}, takes a significant step forward by employing anatomy and pathology masks to guide X-ray generation. This ensures the precise placement of pathologies while maintaining organ-level accuracy. However, the reliance on detailed pathology masks introduces a dependency on expert input, which may limit scalability when using only textual CXR reports.

Our guided image generation employs binary lung segmentation masks and medical text prompts to validate the contextual fidelity of prompts by ensuring anatomical accuracy and consistent localization of pathologies. The generated images aim to validate the output of earlier phrase grounding modules in our pipeline, by quantifying the consistency between the original image and its synthesized counterpart in the region of interest.

% By integrating these components, the interconnected pipeline ensures mutual validation across modules, enhancing trustworthiness in the AI pipeline and addressing the limitations existing in these modules.

\section{Methods}
\label{sec:methods}
\begin{figure}[ht]
    \centering
    \includegraphics[width=0.8\textwidth]{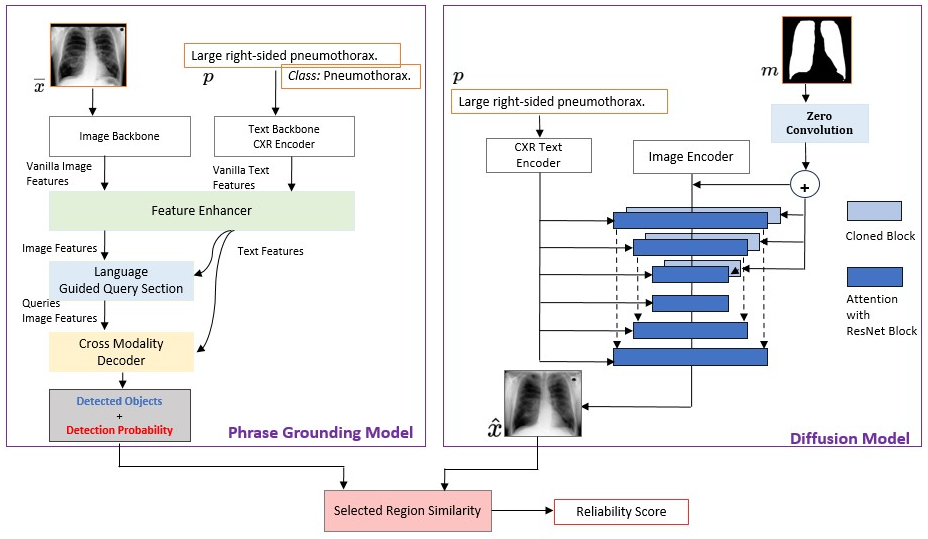}
    \caption{Overview of the VICCA model pipeline: It comprises one primary model (phrase grounding) and one auxiliary model (Diffusion Model). The pipeline takes three inputs: the image ($\overline{x}$), pathology text ($p$), and a binary mask ($m$). The output includes a localization accuracy and reliability score for the pathology text. The output of the Diffusion Model is represented as $\hat{x}$, which should closely approximate the original input image ($\overline{x} \approx \hat{x}$). The ``zero convolution" is 1$\times$1 convolution with both weight and bias initialized to zeros. It serves as a way to integrate the encoded text with the encoded guided binary mask, ensuring the output maintains anatomical structural fidelity.}
    \label{fig:pipeline}
\end{figure}

In this research, we design a multimodal pipeline consisting of two key modules, as shown in Figure \ref{fig:pipeline}. Our primary method involves localizing the input text prompt within CXR images using medical phrase grounding techniques. This step identifies regions of interest (ROIs) corresponding to the described pathologies, forming the foundation for assessing the alignment between textual and visual data.
Subsequently, the remaining components focus on validating the semantic content and reliability of the prompt by indirectly assessing features within the generated image.
We design an auxiliary module, a diffusion model, capable of generating synthetic CXR images based on the input text prompt. This model ensures fidelity to the anatomical structure of the original CXR image while accurately reflecting the prompt's content into visual features. These features are used to support and validate the semantic meaning of the prompt. 
Using the phrase grounding model, we obtain bounding box localizations of the described pathologies. We then leverage the generation model to cross-verify whether the semantic content of the prompt is visually align within these localized regions of bounding boxes.
By generating images with pathologies similar to those described in the text, this approach facilitates a detailed comparison between the original and synthetic CXRs, allowing for a robust evaluation of the consistency between text and image pathologies.

Ultimately, this pipeline produces a comprehensive set of outputs, both visual and textual, accompanied by dual scores, one assessing visual accuracy and the other evaluating semantic alignment. These scores provide critical insights into the reliability of the report, aiding specialists in their decision-making process.

\subsection{Chest X-ray Phrase Grounding}

Our goal in using a phrase grounding model specialized for chest X-ray data is to localize pathology features described in the input text report to their corresponding regions on the image, providing a visual representation as the first step in the interpretability process.
We employ Grounding Dino \cite{42liu2023groundingdino}, a state-of-the-art model designed for phrase grounding in natural images. This model excels in associating textual descriptions with specific regions of interest, leveraging its advanced architecture for precise localization.

\subsubsection{Grounding Dino for Natural Images}

The Grounding DINO architecture is designed for grounded object detection, integrating visual and textual understanding. It begins with a backbone, such as a ResNet or Swin Transformer, which extracts multi-scale features from the input image. These features are then processed by a transformer encoder, employing multi-head self-attention to capture global contextual relationships across the image.

A transformer decoder takes the encoded features and object queries, which are learnable embeddings, to decode information for predicting object regions and their alignment with textual descriptions. A separate text encoder, often derived from a pretrained language model like BERT \cite{BertDevlin2019BERTPO}, processes the input text query, embedding it into a feature space aligned with the visual representations.

The model employs a multimodal alignment to ensure that the embeddings of corresponding text and image elements share a unified feature space, typically using a contrastive learning approach. Predictions are made through specialized heads: one for bounding box localization and another for scoring the alignment between textual queries and detected objects. The training process uses a Hungarian matching loss for object-query assignments, a box regression loss for refining bounding boxes, and a grounding loss to enforce accurate text-object associations. This architecture enables robust performance in tasks requiring simultaneous visual and textual reasoning.

\begin{figure}
    \centering
    \includegraphics[width=0.5\linewidth]{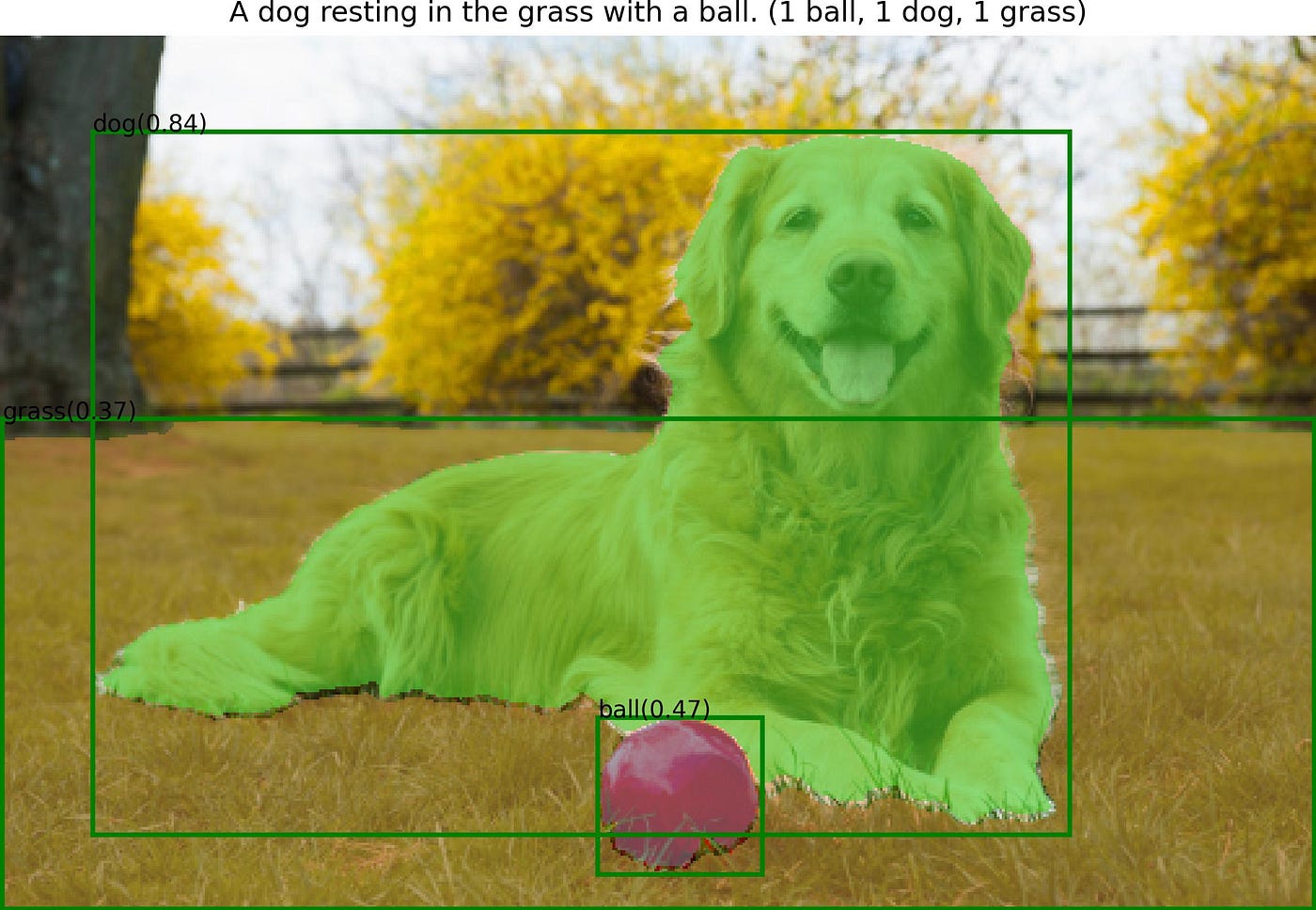}
    \caption{An example of phrase grounding using the Grounding Dino model \cite{42liu2023groundingdino} on a natural image. The output includes localization bounding boxes for all objects mentioned in the text that are present in the image, along with their associated detection probabilities.}
    \label{fig:gd_example}
\end{figure}

Figure \ref{fig:gd_example} demonstrates an example of using the Grounding Dino model for the phrase grounding task. The model processes a natural language prompt (``A dog resting in the grass with a ball") and localizes the described objects (one dog, one patch of grass, and one ball) within the associated image, generating corresponding bounding boxes for each identified region.

\subsubsection{Adapting Grounding Dino to Medical Contexts}
\begin{figure}[t!]
    \centering
    \includegraphics[width=0.5\linewidth]{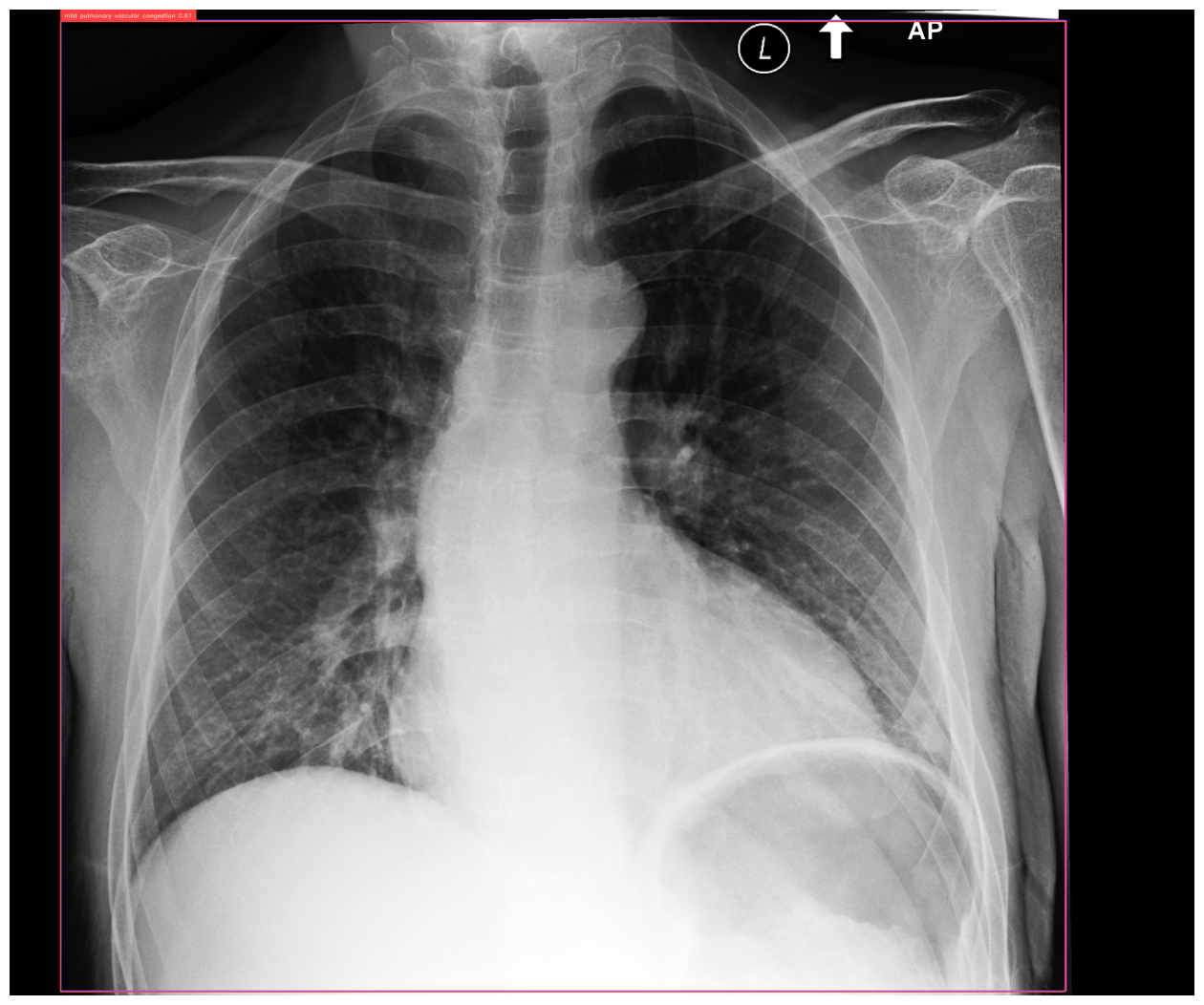}
    \caption{An example of phrase grounding using the Grounding Dino model trained exclusively on natural data with the phrase ``Cardiomegaly with mild pulmonary vascular congestion". The model inaccurately localizes the entire image as the detected object.}
    \label{fig:gd_cxr_annot}
\end{figure}

While Grounding Dino is highly effective for natural images, its direct application to medical imaging is challenging due to the unique demands of the domain. Medical phrase grounding requires a precise spatial alignment and the ability to interpret subtle pathological features, which are often absent in natural datasets. Furthermore, adapting such models necessitates fine-tuning with annotated medical data, including bounding boxes for pathologies and detailed text descriptions.
Figure \ref{fig:gd_cxr_annot} presents a chest X-ray image grounded to the phrase ``Cardiomegaly with mild pulmonary vascular congestion" using the Grounding Dino model trained on natural data. Despite the simplicity of the medical prompt and the accessibility of the text features for even non-experts, the model incorrectly identifies the entire image as the object. This highlights the challenge of capturing subtle pathological features in medical imaging, underscoring the need for domain-specific adaptations.

\begin{figure}[t!]
        \centering
        \includegraphics[width=1\linewidth]{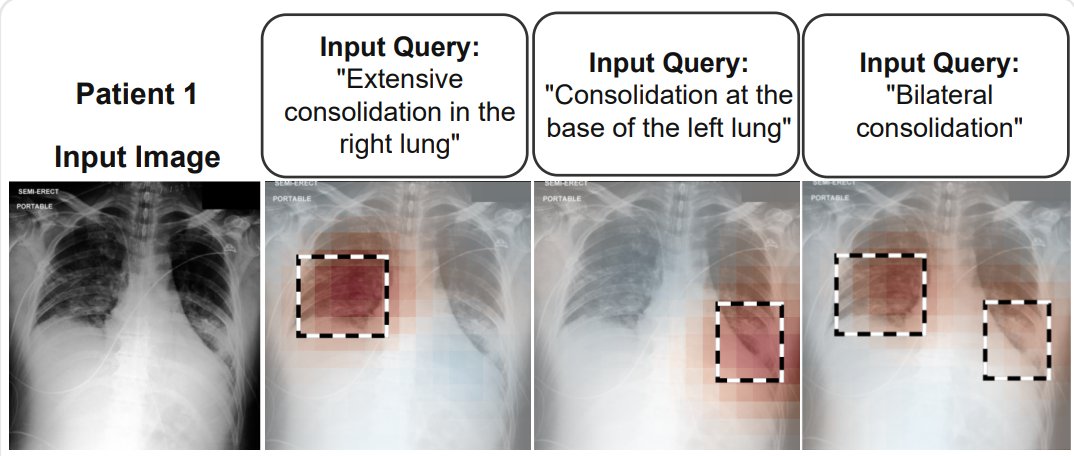}
        \caption{A well-annotated dataset for phrase grounding, featuring bounding box locations, associated text queries, and corresponding pathology descriptions \cite{boecking2022making}.}
        \label{fig:ms-cxr}
    \end{figure}

We fine-tuned Grounding Dino on CXR datasets to overcome these challenges, customizing it for medical applications. For this process, we used two key datasets:

\paragraph{1. MS-CXR Dataset \cite{boecking2022making}}
A specialized dataset designed for phrase grounding tasks, as illustrated in Figure \ref{fig:ms-cxr}. It contains 1,162 annotated CXR images, each paired with detailed bounding box annotations and corresponding pathology descriptions, making it a valuable resource for developing and evaluating phrase grounding models. However, the dataset's limited size poses a challenge, as it is insufficient for training a conventional object detection model. This necessitates strategies such as few-shot learning or the incorporation of additional datasets to ensure robust model performance.

\paragraph{2. VinDr-CXR Dataset \cite{VinDrCXRnguyen2022vindrcxropendatasetchest}} 
A larger dataset with over 18,000 subjects' annotated images, providing bounding box locations for 17 common thoracic diseases. 
However, VinDR-CXR lacks detailed written reports associated with each subject, which is critical for our phrase grounding purposes. 

One potential solution to this issue is the use of automated report generation models to create descriptive reports. However, as our goal is to enhance the interpretability of such models, we opted not to include them during training to avoid introducing potential biases.
Instead, we enhanced the dataset by associating pathologies with their anatomical regions to enrich the spatial information.

To support this process, we leveraged the chest ImaGenome dataset \cite{79imagenomeWu2021ChestID}, which provides detailed, structured annotations of CXR images through a radiologist-constructed ontology. Each CXR is represented as an anatomy-centered scene graph, connecting 36 anatomical locations with bounding boxes and their associated attributes. The dataset includes over 1,256 types of relation annotations between anatomical structures, more than 670,000 localized comparisons across sequential exams (e.g., improved, worsened, or stable), and a manually curated gold-standard scene graph set for 500 patients.

To associate the VinDr-CXR pathologies with precise anatomical regions, we developed an automatic detection method based on the DETR (\textbf{Detection Transformer}) model \cite{detr}, a state-of-the-art object detection framework.
DETR combines a convolutional neural network backbone with a transformer architecture, offering several advantages, such as its ability to generate unique predictions without relying on traditional post-processing steps like non-maximum suppression. These properties make DETR particularly suited for detecting overlapping and closely spaced regions, which are common in medical imaging tasks.

\begin{figure}
    \centering
    \includegraphics[width=0.5\textwidth]{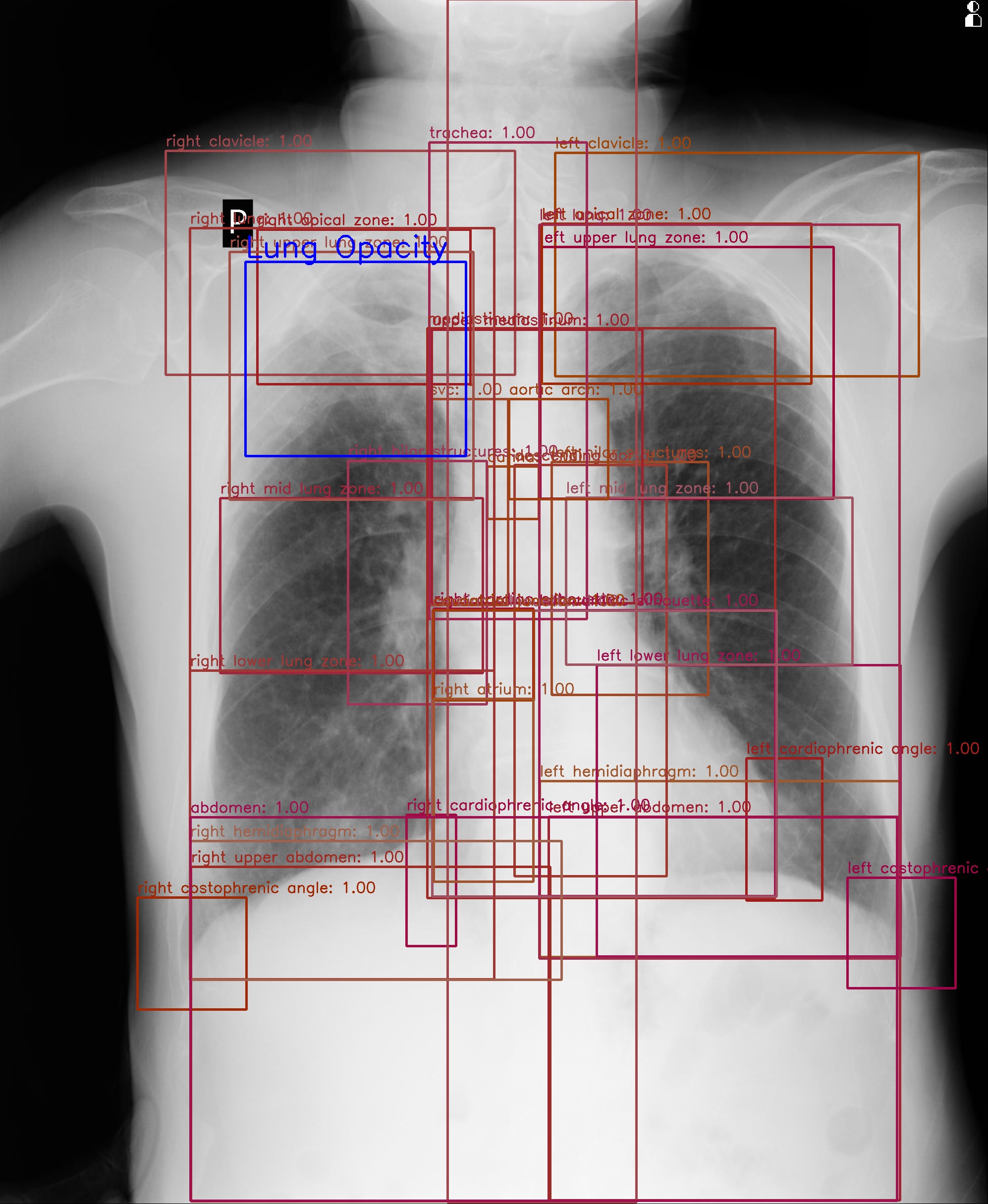}
    \caption{An example CXR image from the VinDr-CXR dataset. The blue bounding box represents the dataset-provided annotation for the pathology present in the image. The red bounding boxes indicate the anatomical regions automatically detected by our trained model.}
    \label{fig:detr_vindr}
\end{figure}

While the original DETR model was pretrained on natural images, its transformer based architecture and end-to-end training pipeline make it adaptable to complex medical datasets. For our study, we trained DETR from scratch on the chest ImaGenome dataset, leveraging its high-quality bounding box annotations for anatomical regions. This adaptation benefits from DETR's global attention mechanism, enabling precise identification of anatomical structures, even when these regions are subtle or overlap significantly.
An example output is shown in Figure \ref{fig:detr_vindr}, where bounding boxes in red represent automatically detected anatomical regions.

Since the VinDr-CXR dataset provides pathology labels and their corresponding bounding boxes but lacks anatomical descriptors, we enriched the dataset by associating each pathology label with the closest anatomical region. This was achieved using empirical distance and intersection over union (IoU) metrics. 
For instance, in the case of Figure \ref{fig:detr_vindr}, a bounding box labeled ``Lung Opacity" that overlaps most with the ``right upper lung zone" was annotated as ``Lung Opacity in right upper lung zone". This refinement significantly improves the granularity and spatial context of the dataset, making it more suitable for tasks requiring precise alignment between text and image features.
Moreover, this approach can be considered as a frugal model, leveraging existing datasets and computational resources efficiently while minimizing dependency on extensive new data annotations. Such a framework can enhance a wide range of medical imaging tasks, including detection, segmentation, and classification. 

Given that this research does not involve direct radiologist feedback, we manually verified 200 samples by referencing existing radiology descriptors \footnote{\url{www.radiologymasterclass.co.uk}} to ensure the accurate association of pathologies with anatomical regions. Based on this validation, we recalibrated the automated augmentation process for the remaining dataset, ensuring consistency and reliability across all annotations.
For instance, our initial approach primarily assigned each pathology to the closest and most overlapped anatomical region. However, in cases of bilateral pathologies, this method could result in one side being excluded. To address this, we refined the model to account for bilateral occurrences by considering both left and right regions when they overlap with the pathology's bounding box.

The fine-tuning process involved few-shot training, using a subset of annotated images to adapt the model's weights to the medical domain while preserving its pretrained capabilities for natural data. This step is critical to balance generalization and domain-specific accuracy.

\subsubsection{Model Architecture and Fine-Tuning}
As shown in Figure \ref{fig:gdmodel}, we fine-tuned the Grounding Dino's architecture in the following components:

\begin{figure}
    \centering
    \includegraphics[width=\textwidth]{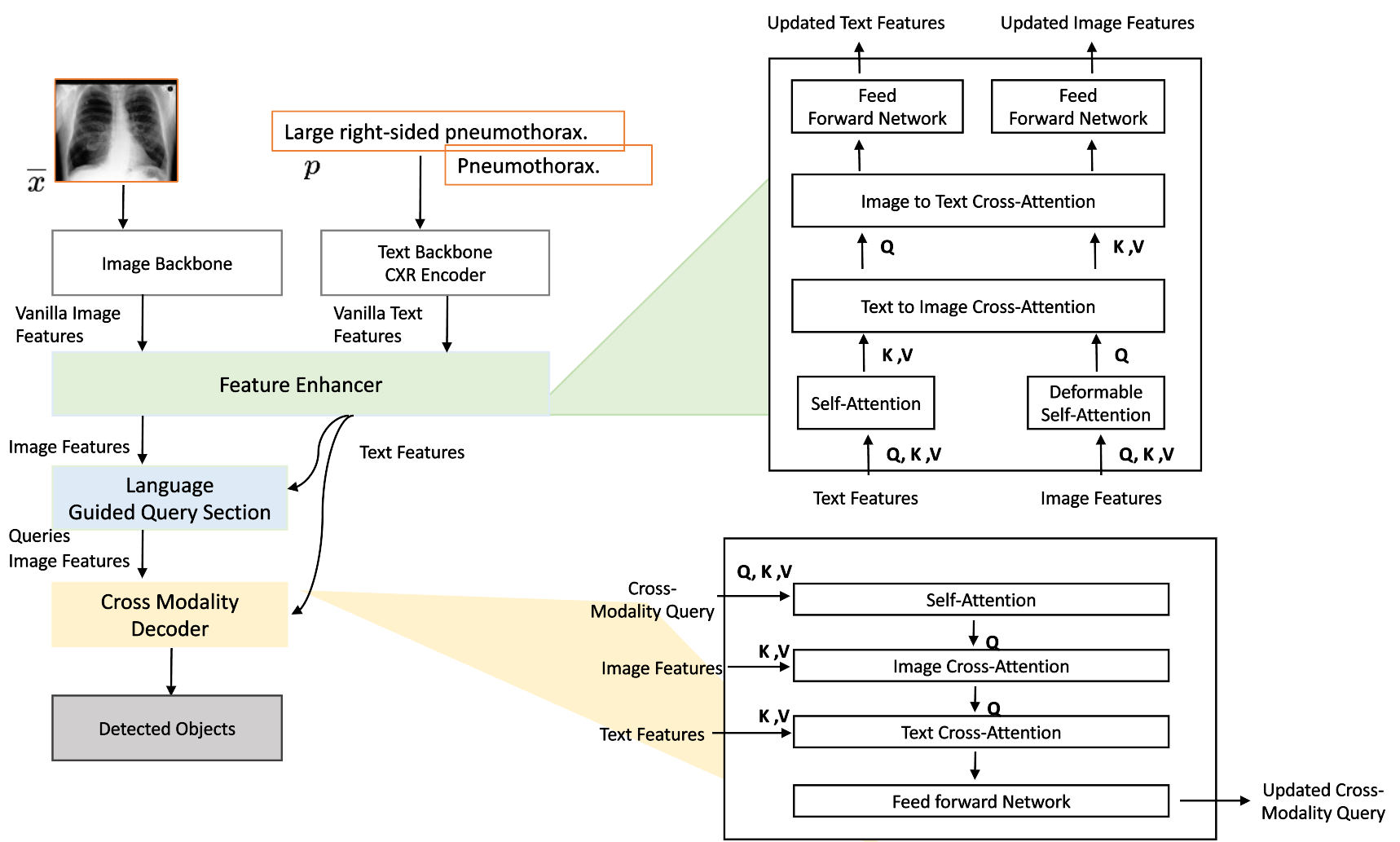}
    \caption{The architecture of the Grounding DINO model integrates the BiomedVLP-CXR-BERT as the text encoder for medical text attention. This model grounds the provided text input onto the image using cross-attention maps between the image and text.}
    \label{fig:gdmodel}
\end{figure}

\begin{itemize}
    \item \textbf{Language Encoder} tokenizes and embeds textual prompts for semantic representation. 
    Grounding Dino uses BERT encoder \cite{BertDevlin2019BERTPO}, a text encoder pretrained on the English language using a masked language modeling (MLM) objective, empirically proven powerful for natural language tasks, equipped with a large token vocabulary. However, numerous tokens in this vocabulary are not relevant to the medical domain. Despite efforts to integrate models such as ClinicalBERT \cite{clinicalbert}, specifically trained for medical applications, into Grounding DINO's training process, this did not result in notable enhancements in feature extraction. To overcome these limitations, we integrated BiomedVLP-CXR-BERT \cite{boecking2022making}, a text encoder specifically trained on chest X-ray reports. By leveraging BiomedVLP-CXR-BERT in the fine-tuning process, we aligned the text encoder's outputs more closely with the medical context, significantly enhancing localization outcomes.
    
    By integrating the BiomedVLP-CXR-BERT encoder into the Grounding DINO model, we effectively transformed the previously natural language outputs into medically contextualized ones. This adjustment resulted in a substantial performance enhancement. Specifically, the general DETR-like loss during training decreased from 35.85 when using the BERT encoder to 8.79 when using the BiomedVLP-CXR-BERT encoder. 
    Figure \ref{fig:training_loss} compares the training loss progression between the two encoders, showing that the specialized encoder leads to both lower loss and faster convergence.
    The model using the BERT encoder shows significant fluctuations in training loss, ranging from 30.08 to 44.29. This high variation is caused from unstable feature matching in the phrase grounding model, indicating that the BERT encoder struggles to extract clinically relevant features from the text. As a result, it fails to consistently align textual and visual representations, which disrupts the learning process.
    
    To ensure that overfitting is not occurring, Figure \ref{fig:training_loss_zoom} provides a zoomed-out view of the training curve with BiomedVLP-CXR-BERT. Despite fluctuations, the loss remains within a stable range [7.6, 11.8], which is consistent with the typical behavior of Grounding DINO models \cite{ferreira2025beyond}, before converging to 8.79 by epoch 407. This pattern suggests normal variability during training rather than evidence of overfitting. Validation loss curves are omitted, as the objective of Figure \ref{fig:training_loss_genral} is to contrast the training dynamics of the different encoders rather than to evaluate generalization performance directly. Model generalization is more appropriately assessed through the metrics reported in Sections \ref{subsec:PGCXR_Val} and \ref{subsec:pipline_eval}.

    \begin{figure}[t!]
    \centering
    \begin{subfigure}[t]{0.5\textwidth}
        \centering
        \includegraphics[width=\linewidth]{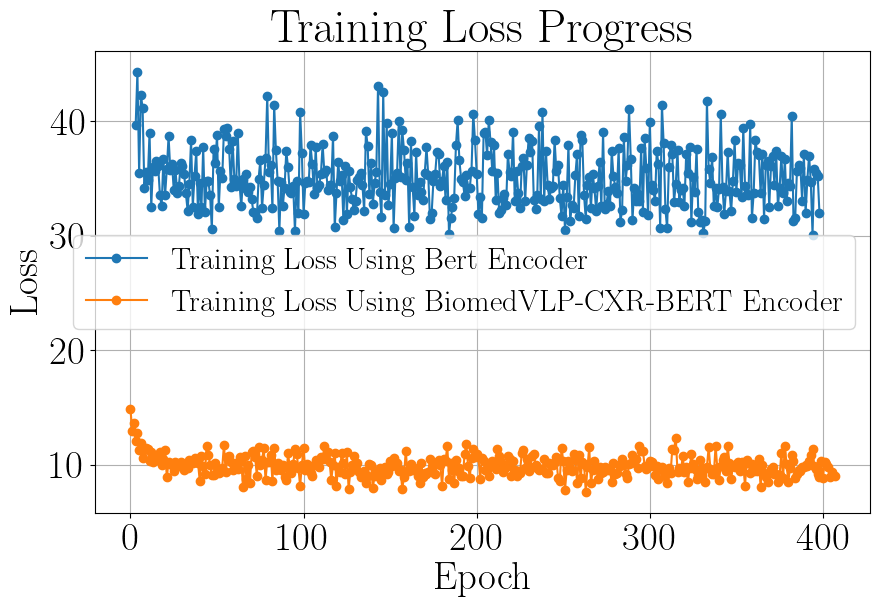}
        \caption{}
        \label{fig:training_loss}
    \end{subfigure}%
    ~ 
    \begin{subfigure}[t]{0.5\textwidth}
        \centering
        \includegraphics[width=\linewidth]{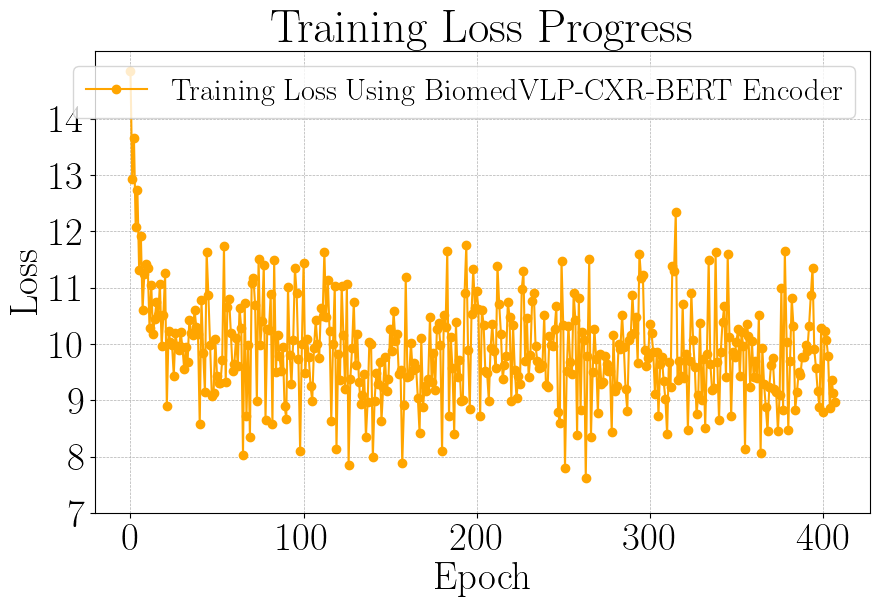}
        \caption{}
        \label{fig:training_loss_zoom}
    \end{subfigure}
    \caption{(a) Training loss progression for the phrase grounding model using two text encoders (BERT and BiomedVLP-CXR-BERT). The specialized BiomedVLP-CXR-BERT encoder (orange curve), designed for CXR text, significantly reduces the loss and achieves faster convergence compared to using the BERT encoder (blue curve). (b) A zoomed-out view of the loss curve for the BiomedVLP-CXR-BERT encoder shows a stable yet non-monotonic pattern, suggesting variability in learning but no clear signs of overfitting.}
    \label{fig:training_loss_genral}
\end{figure}
    
    \item \textbf{Image encoder} processes input images to extract relevant visual features. Grounding Dino supports both ResNet \cite{ResNet} and Swin Transformer \cite{liu2021Swin} architectures as backbones. For our fine-tuning, we employed only the Swin Transformer due to its superior performance on high-resolution images and its patch-merging strategy, which effectively aligns text and image features both globally and locally. This choice proved particularly advantageous for the intricate spatial details present in chest X-ray analysis.

    \item \textbf{Loss Function:} We adopted the original loss configuration of Grounding Dino, which includes L1 loss and Generalized Intersection over Union (GIoU) for bounding box regression \cite{87zhang2022glipv}. The model also employs a contrastive loss to associate predicted objects with their corresponding classifications. This process involves computing the dot product between each query and text feature, generating logits for each text token, and applying focal loss \cite{Lin2017FocalLF} to these logits.
    For fine-tuning, a multi-task loss function combining localization loss (L1 and GIoU) and classification loss (focal loss) was utilized. 
    Training was conducted over 400 epochs, with hyperparameters such as learning rate and batch size optimized via grid search. The final model exhibited a stable loss convergence with the loss stabilizing around 8.7, ensuring both precise localization and robust classification capabilities.
\end{itemize}

Success in the phrase grounding task provides the first score in our dual-scoring mechanism, offering a detection accuracy metric for each bounding box. Table \ref{tab:result_gd} presents a case study using the optimized Grounding DINO model with two different textual inputs. The CXR image was annotated with the accurate diagnosis of ``Cardiomegaly with mild pulmonary vascular congestion," achieving a bounding box detection accuracy of 76\%. Conversely when exposed to an unrelated prompt, ``Left-sided Pulmonary Edema," the accuracy dropped significantly to 20\%, localizing an anomaly in the left lung but failing to identify edema.
This result highlights the model's capability to differentiate and accurately localize the correct pathology when provided with a relevant text prompt, while demonstrating reduced performance for unrelated or incorrect prompts.

\begin{table}[]
    \centering
    \caption{The results of the phrase grounding model using two different text prompts. The ground truth report, ``Cardiomegaly with mild pulmonary vascular congestion," got the bounding box detection with 76\% accuracy. The last column shows the results with a random prompt, ``Left-sided Pulmonary Edema," which achieved a bounding box accuracy of 20\%.}
    \begin{tabular}{|l|p{3.1cm}|p{3.1cm}|}
    \hline
        \rotatebox{90}{Input    } & Cardiomegaly with mild pulmonary vascular congestion. & Left-sided Pulmonary Edema. \\
        \hline
        \rotatebox{90}{Output}
        &
            % \centering
            \includegraphics[width=0.2\textwidth]{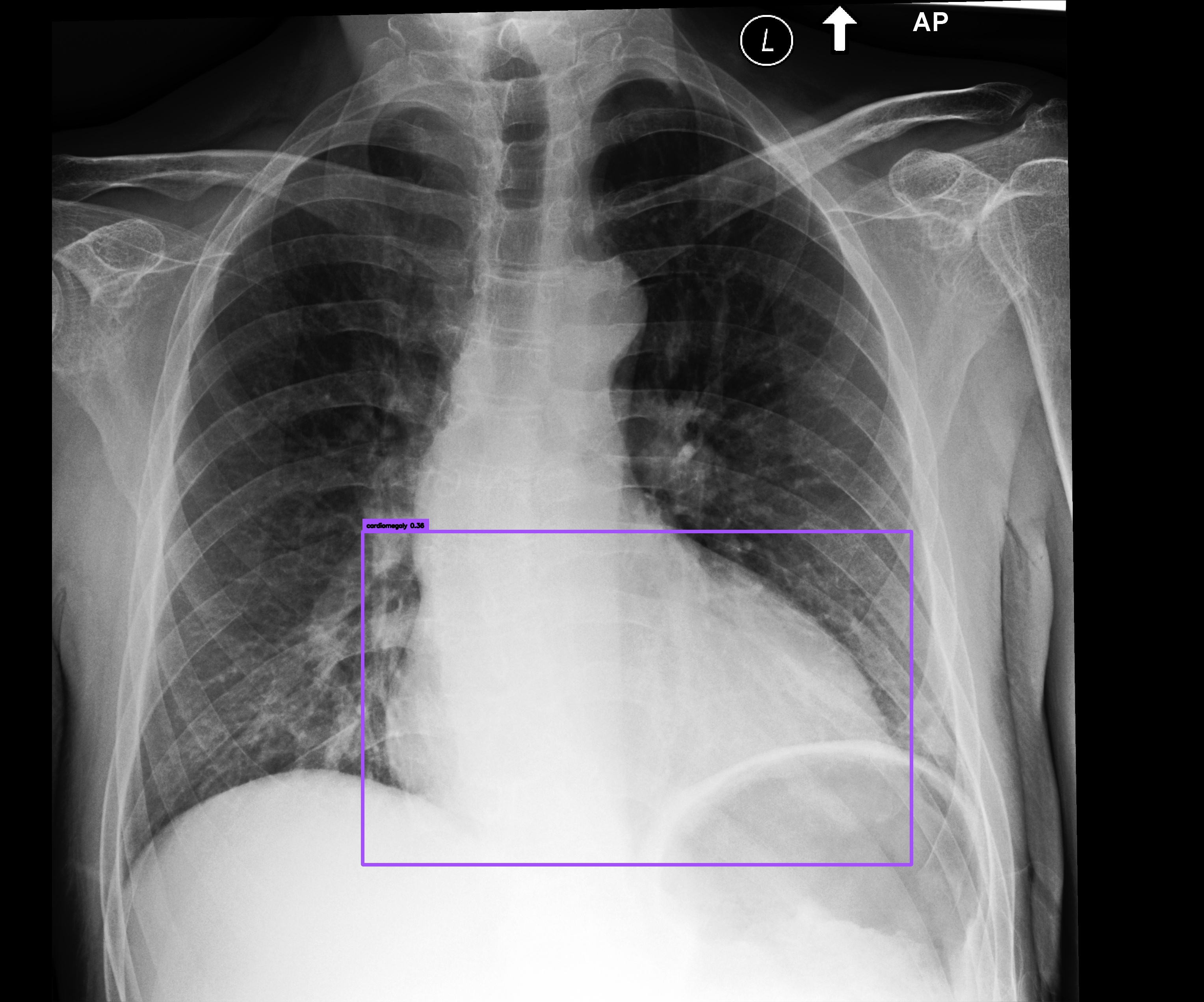}
            % \caption{Caption}
            \label{fig:sampleGD1}
         & 
            % \centering
            \includegraphics[width=0.2\textwidth]{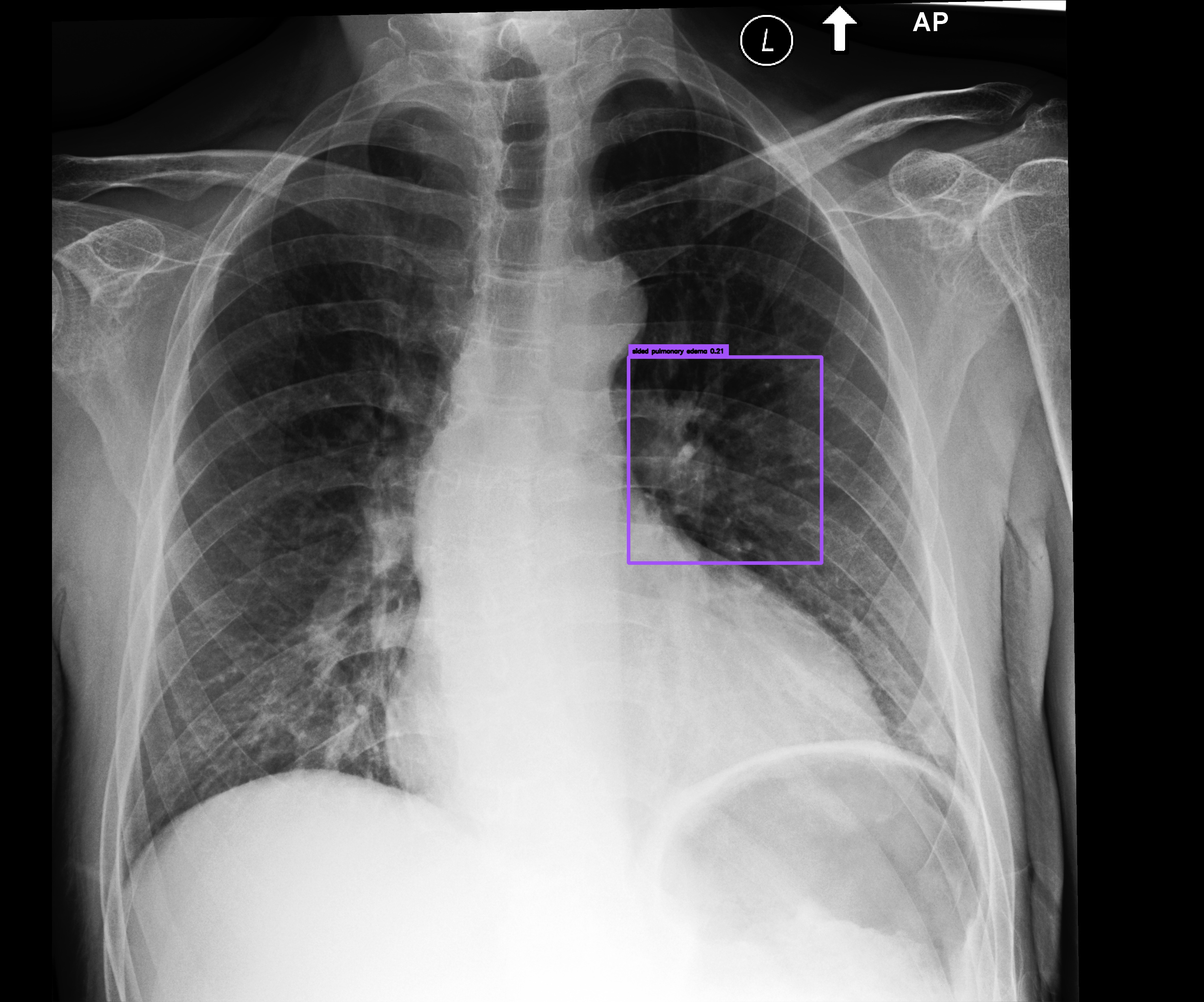}
            % \caption{Caption}
            \label{fig:sampleGD2}\\
            \hline
    \end{tabular}
    \label{tab:result_gd}
\end{table}

While our enhancements improve the visual interpretability of the text report, the model's outputs still require radiologist validation to ensure reliability and comprehensive accuracy. To address this limitation and achieve independence from expert feedback in this research, we introduce an auxiliary model specifically designed to evaluate the reliability and accuracy of the bounding box localizations.

To ensure a proper alignment between the text prompt and the image content, where the text accurately describes the image and the image features validate the prompt's meaning, we introduce a reliability score. This score quantifies the alignment, providing an objective measure to assess the trustworthiness of the model's outputs.

\subsection{\textit{Auxiliary Model:} Stable Diffusion for CXR Generation with Anatomy Preservation}

We introduce an auxiliary model to enhance the reliability of the visual interpretation of the text report. Specifically, we utilize a CXR generation model to produce an alternative CXR image that is anatomically similar to the original but guided by the input text report. This approach provides an additional set of CXR images that can be compared to the localized regions in the original image. The comparison allows us to evaluate the report's accuracy and its alignment with the image by determining whether the report provides sufficient spatial details, such as pathology and anatomical regions. If accurate, the localized regions in the original CXR image should closely align with the corresponding areas in the generated images.

\begin{table}[t!]
    \centering
    \begin{tabular}{c|c}
    \hline
    \multicolumn{2}{c}{Generated Caption: A dog frolic in a grassy field, bathed in the warm sunlight.}\\
    \hline
    Reference Image&Generated Image\\
    \hline
     \includegraphics[width=0.45\linewidth]{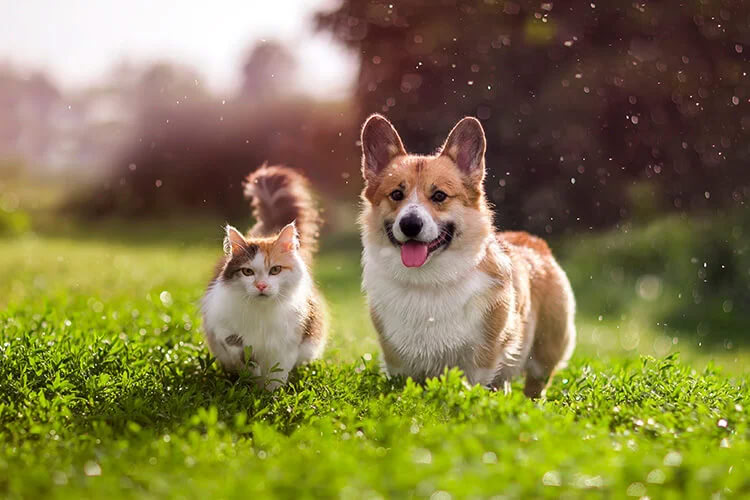}&\includegraphics[trim={0 0 0 5cm}, clip, width=0.4\linewidth]{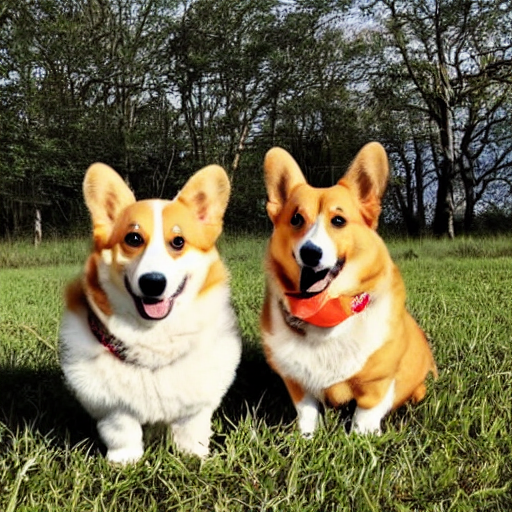}\\
     \hline
     \multicolumn{2}{c}{Phrase Grounding}\\
     \hline
     \includegraphics[width=0.45\linewidth]{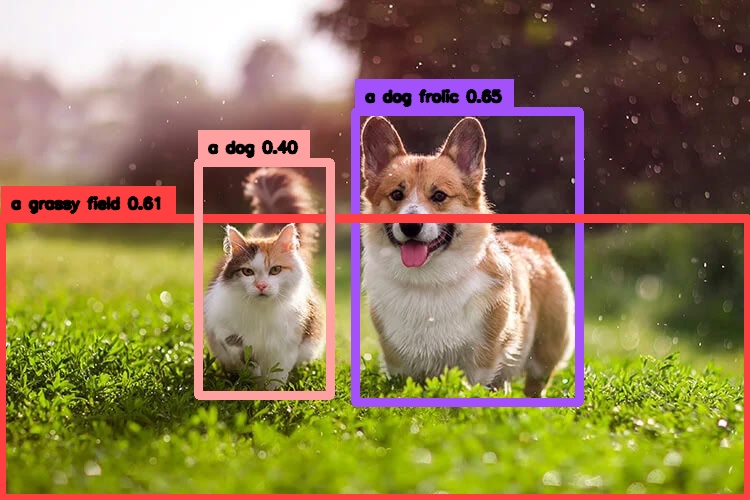}&\includegraphics[trim={0 0 0 4.5cm}, clip, width=0.4\linewidth]{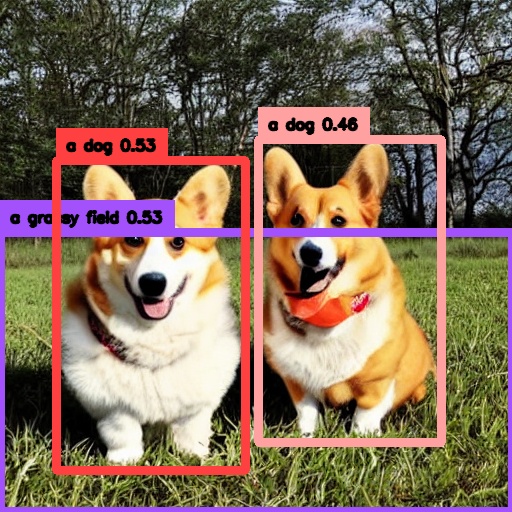}\\
     \hline
    \end{tabular}
    \caption{Comparison between the reference image containing both a cat and a dog (left) and a synthetic image of two dogs generated based on the caption (right). The second row illustrates object localization results using the generated caption, highlighting how the model aligns text references with visual regions in both real and synthesized images.}
    \label{tab:application}
\end{table}

To contextualize this approach, consider an analogous example from the natural image domain. Table \ref{tab:application} presents a simple scenario involving an image containing both a cat and a dog. An AI model generates the caption, ``A dog frolics in a grassy field, bathed in warm sunlight." Notably, the caption omits any mention of the cat. In such cases, detection-based models like Grounding DINO may misclassify or completely overlook the secondary object (in this case, the cat), focusing instead on the explicitly referenced dog.

To resolve this ambiguity, a text-to-image diffusion model can synthesize an image based solely on the given caption. When comparing regions of interest between the original image and the generated one, the dog in the reference image shows a much higher similarity score (64\%) with its synthesized counterpart than the cat does (13\%). This gap suggests that the cat's presence is not semantically supported by the caption, indicating the description may be incomplete or misleading.

Applying this concept to the medical domain, particularly chest X-rays, the 'cat' could represent a missed pathology such as 'pneumonia'. If the report omits pneumonia but includes another condition like 'pleural effusion', the localization model might incorrectly attribute a region indicative of pneumonia with low detection accuracy to pleural effusion.

Although such discrepancies are relatively easy to detect in natural images, they can be far more subtle and clinically significant in medical domains. In such contexts, auxiliary models like ours can play a critical role in revealing omissions, ambiguities, or inconsistencies in machine-generated reports.

This example parallels our medical imaging approach and demonstrates how multimodal validation, linking text and image, can support automated fact-checking, semantic alignment, and trust calibration in AI systems.

Given that most thoracic diseases occur within or near the lung region as visible on chest X-rays, and to ensure the preservation of anatomical structures similar to the original image, we extract the lung segmentation from the original image. This is achieved using an existing model \footnote{\url{https://github.com/IlliaOvcharenko/lung-segmentation.git}}.

Using the binary lung segmentation image and the input text report, we generate a chest X-ray image through a stable diffusion model originally designed for text-to-image synthesis. By incorporating inpainting techniques, we integrate the binary image into the training process to suit our specific purpose.
For this task, we build upon the stable diffusion model (SD1.5), leveraging its pretrained weights to retain its understanding of general tokens. We then adopt the ControlNet model architecture \cite{zhang2023adding_controlnet} as the backbone and integrate a BiomedVLP-CXR-BERT encoder for tokenizing chest X-ray entities. 

The final architecture of the model is illustrated in Figure \ref{fig:pipeline} diffusion model. 
Similar to the stable diffusion model, ControlNet uses an autoencoder architecture for text-to-image generation. However, in ControlNet, additional cloned encoder blocks are introduced to process the binary image, providing control over the inpainting process. While the original encoder blocks learn, the cloned blocks are frozen, and vice versa. A zero-convolution layer is then used to link the cloned blocks with the locked model, enabling smooth integration and enhanced control.

For training, we use the MIMIC-CXR v2.0.0 dataset \cite{14physionetGoldberger2000-si, 30mimic-cxrJohnson2019}, a large public collection of chest radiographs paired with free-text radiology reports. It includes 227,835 studies from 65,079 patients performed at Beth Israel Deaconess Medical Center. Along with images and corresponding reports, the dataset also provides annotations for thoracic diseases, enabling supervised learning for various medical imaging tasks.
We follow the official MIMIC-CXR train/validation/test split, resulting in 1,879 image-report pairs for validation and 3,082 for testing.

The model is fine-tuned to adapt to CXR data effectively. The training is implemented over 25 epochs, achieving a training loss of 0.078 and a validation loss of 0.125, indicating a stable convergence.

\begin{table}[t!]
    \centering
    \caption{The results of the CXR generative model using two different text prompts. The ground truth report is in the third row, and a random CXR report is in the fourth row. Feature similarity is used to calculate the generation's similarity with the Ground Truth (GT) image.}
    \begin{tabular}{|p{3.5cm}|p{3.5cm}|p{3.5cm}|p{3.5cm}|}
    \hline
    \multicolumn{4}{|c|}{Original Image}\\
    \hline
    \includegraphics[width=0.2\textwidth]{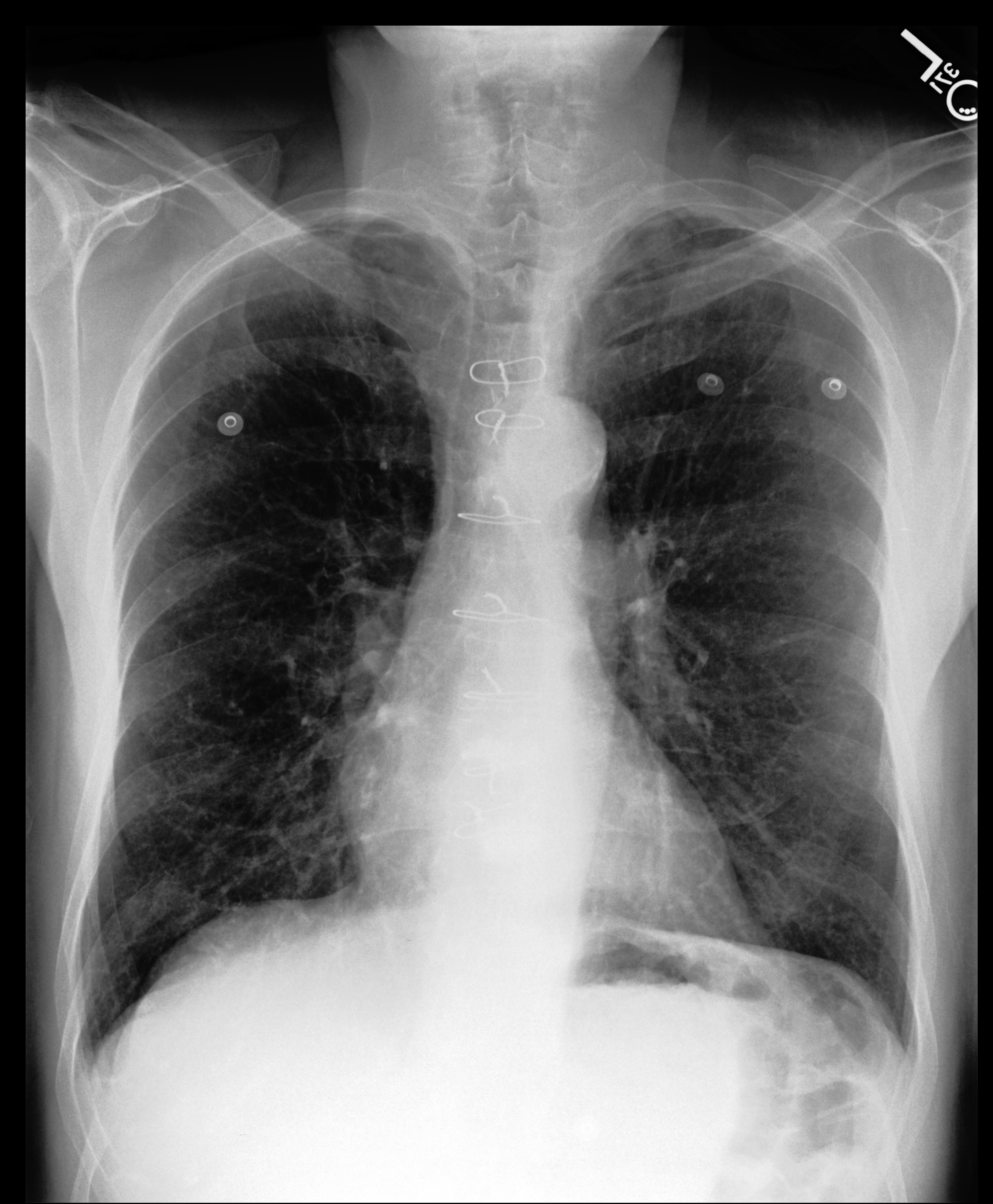} & \includegraphics[width=0.2\textwidth]{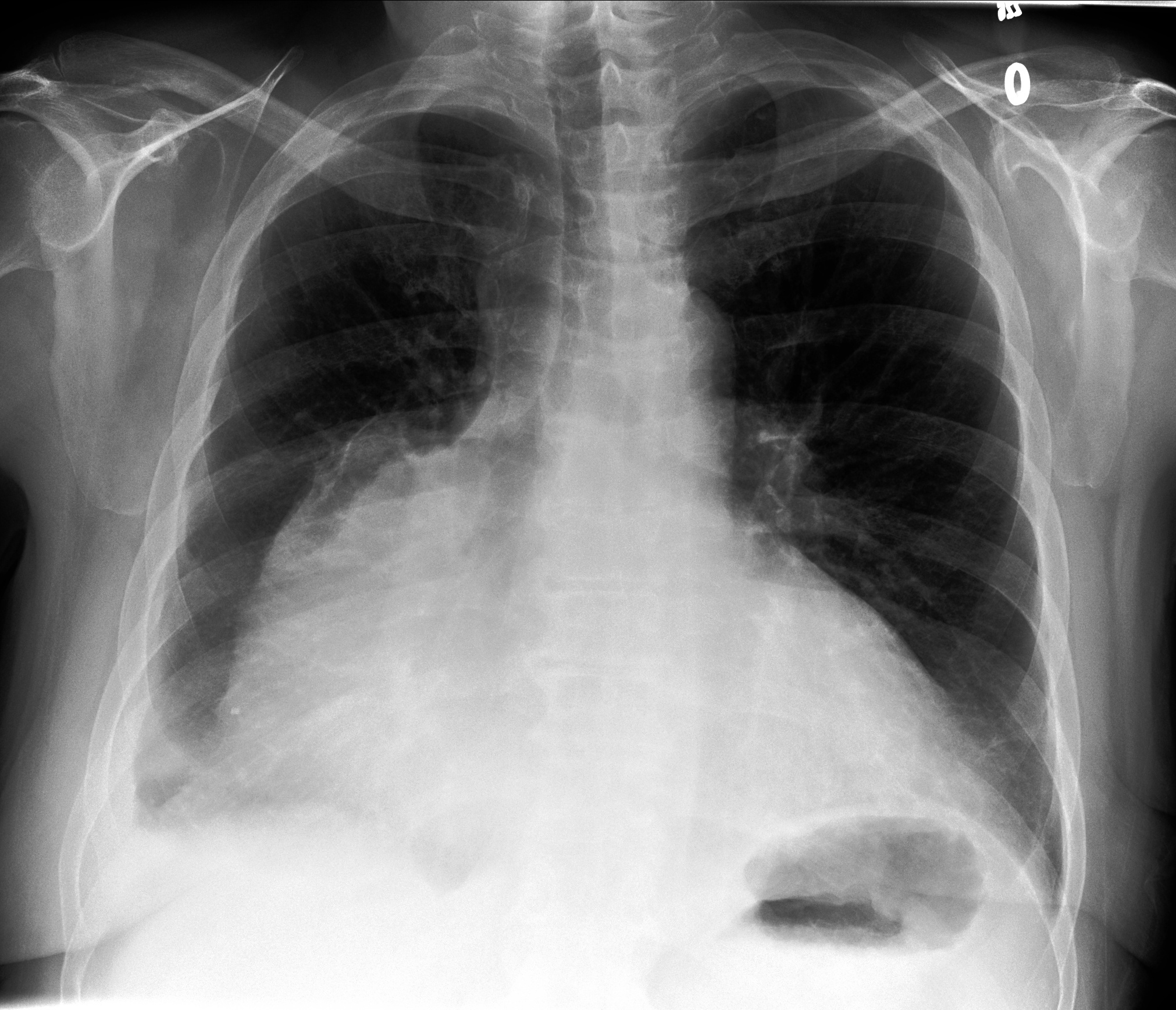} & \includegraphics[width=0.2\textwidth]{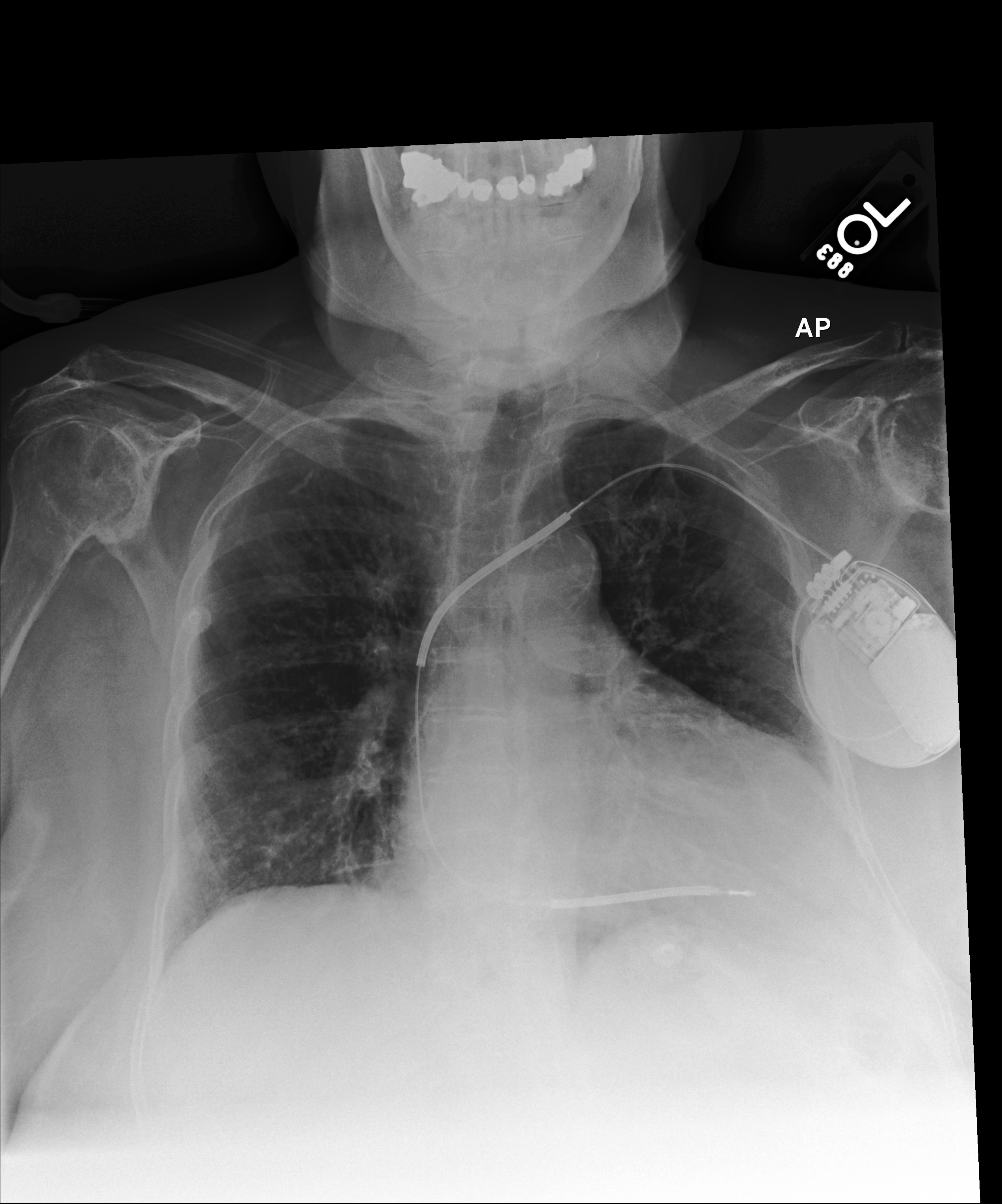} & \includegraphics[width=0.2\textwidth]{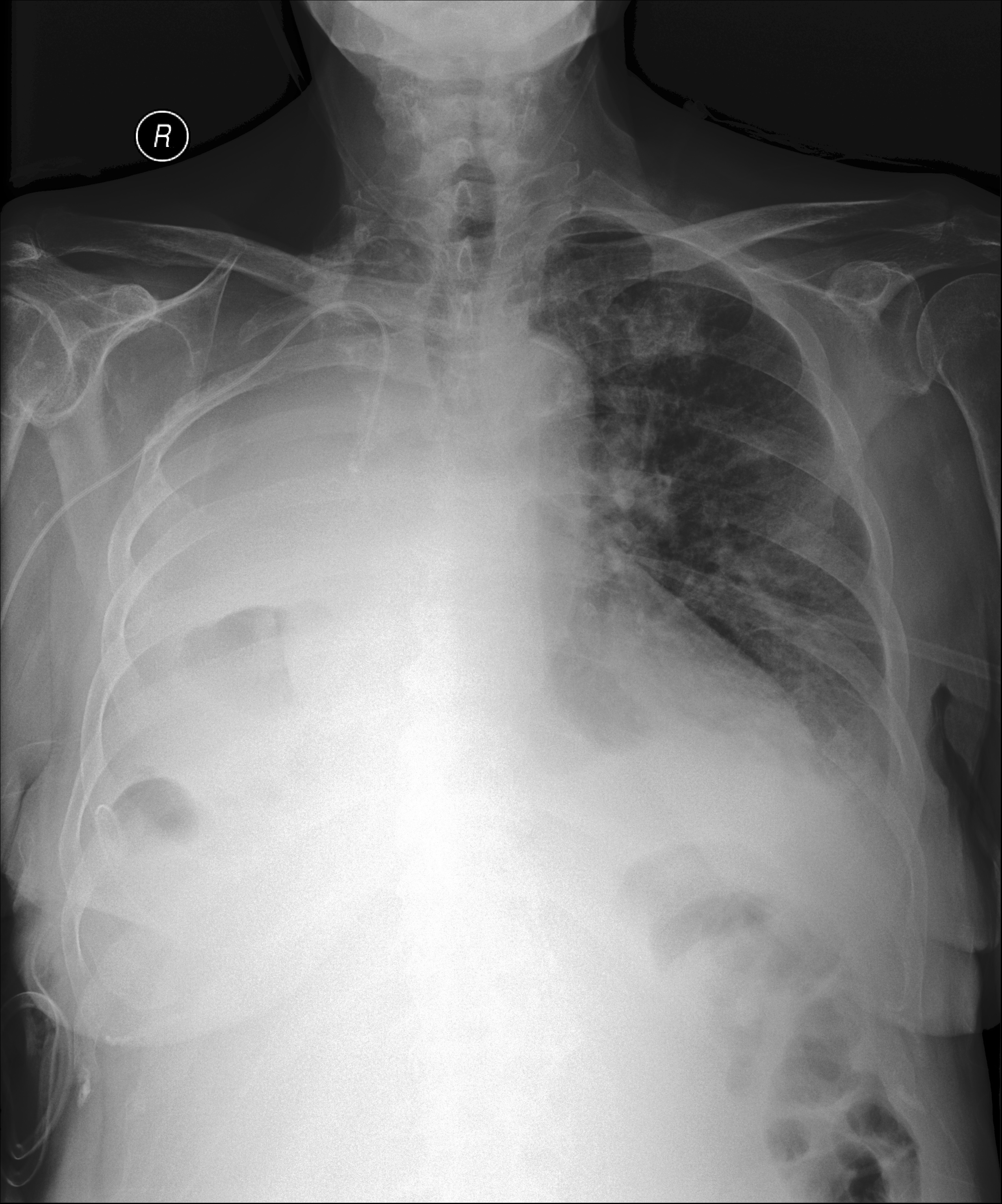} \\
    \hline
    \multicolumn{4}{|c|}{Binary Image Lung Segmentation}\\ 
    \hline
    \includegraphics[width=0.2\textwidth]{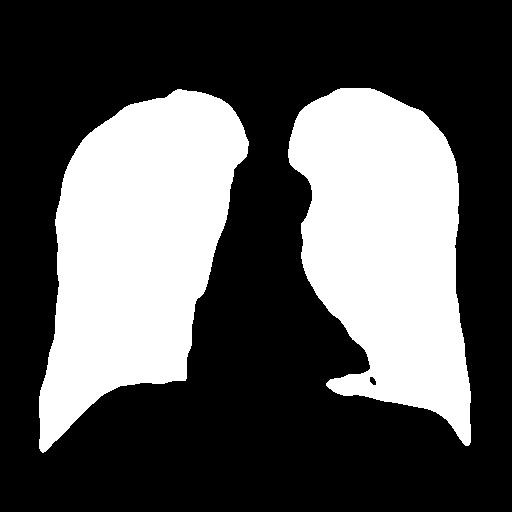} & \includegraphics[width=0.2\textwidth]{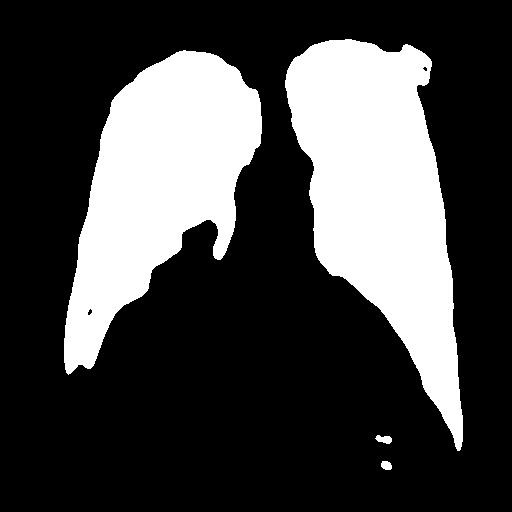} & \includegraphics[width=0.2\textwidth]{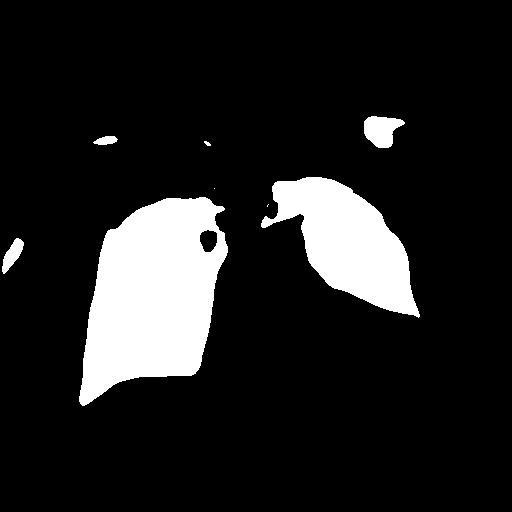} & \includegraphics[width=0.2\textwidth]{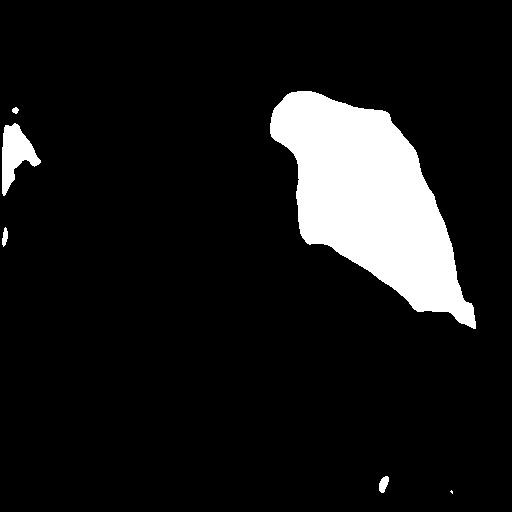} \\
    \hline
    \multicolumn{4}{|c|}{Generated with the Original Report}\\
    \hline
    \includegraphics[width=0.2\textwidth]{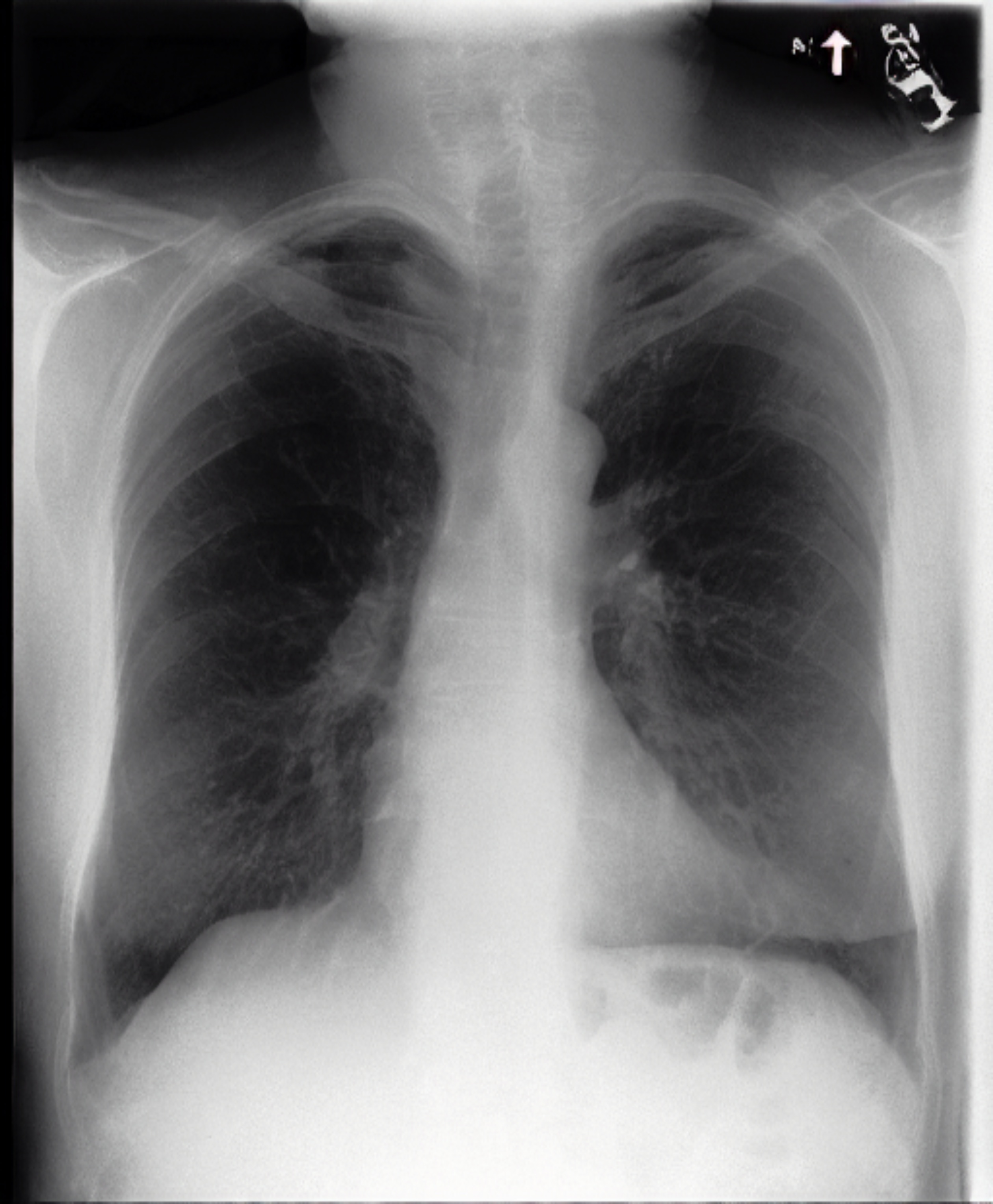} 
    Similarity with the GT: 88.6\%
    & \includegraphics[width=0.2\textwidth]{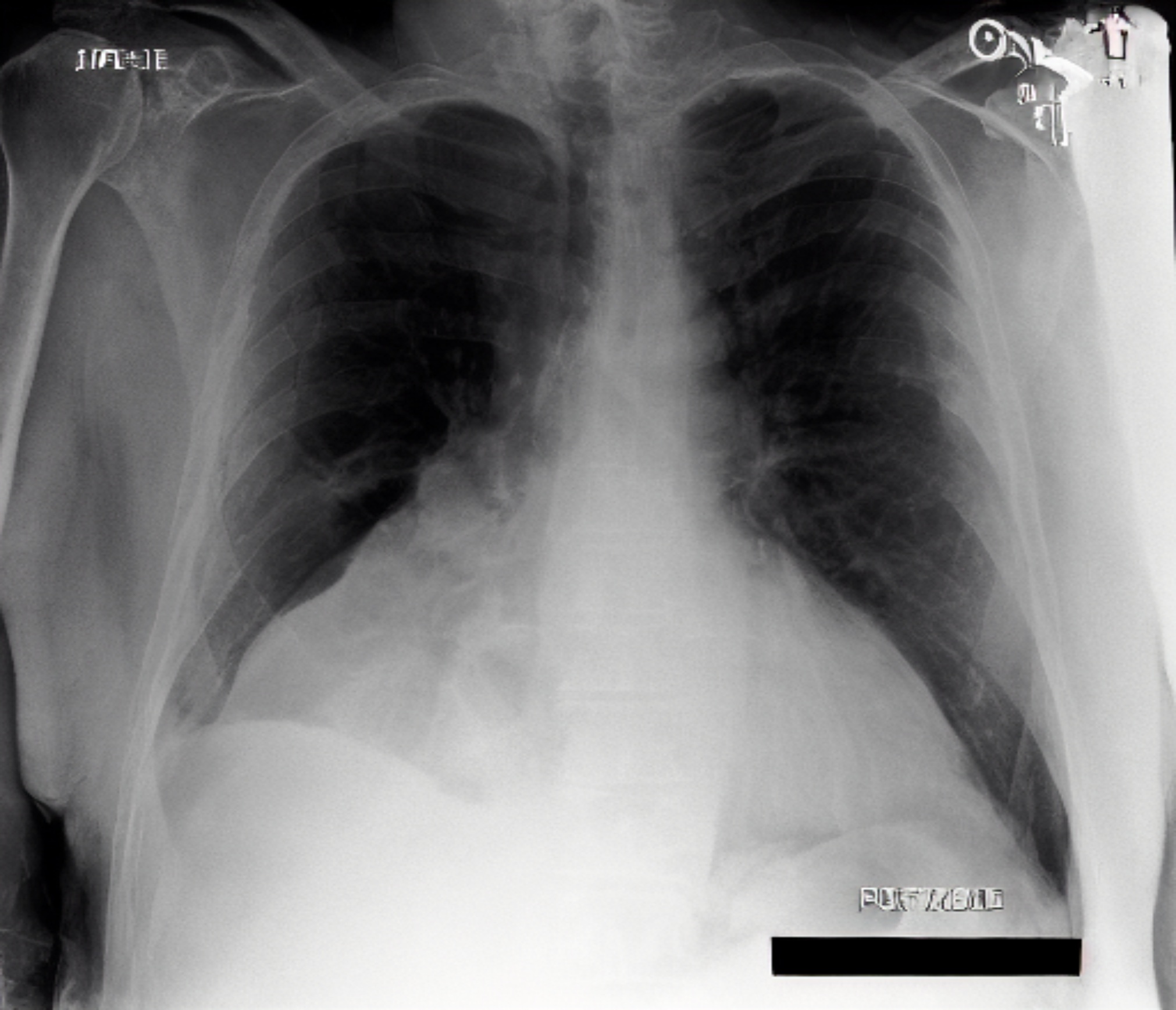} 
    Similarity with the GT: 85.3\%
    & \includegraphics[width=0.2\textwidth]{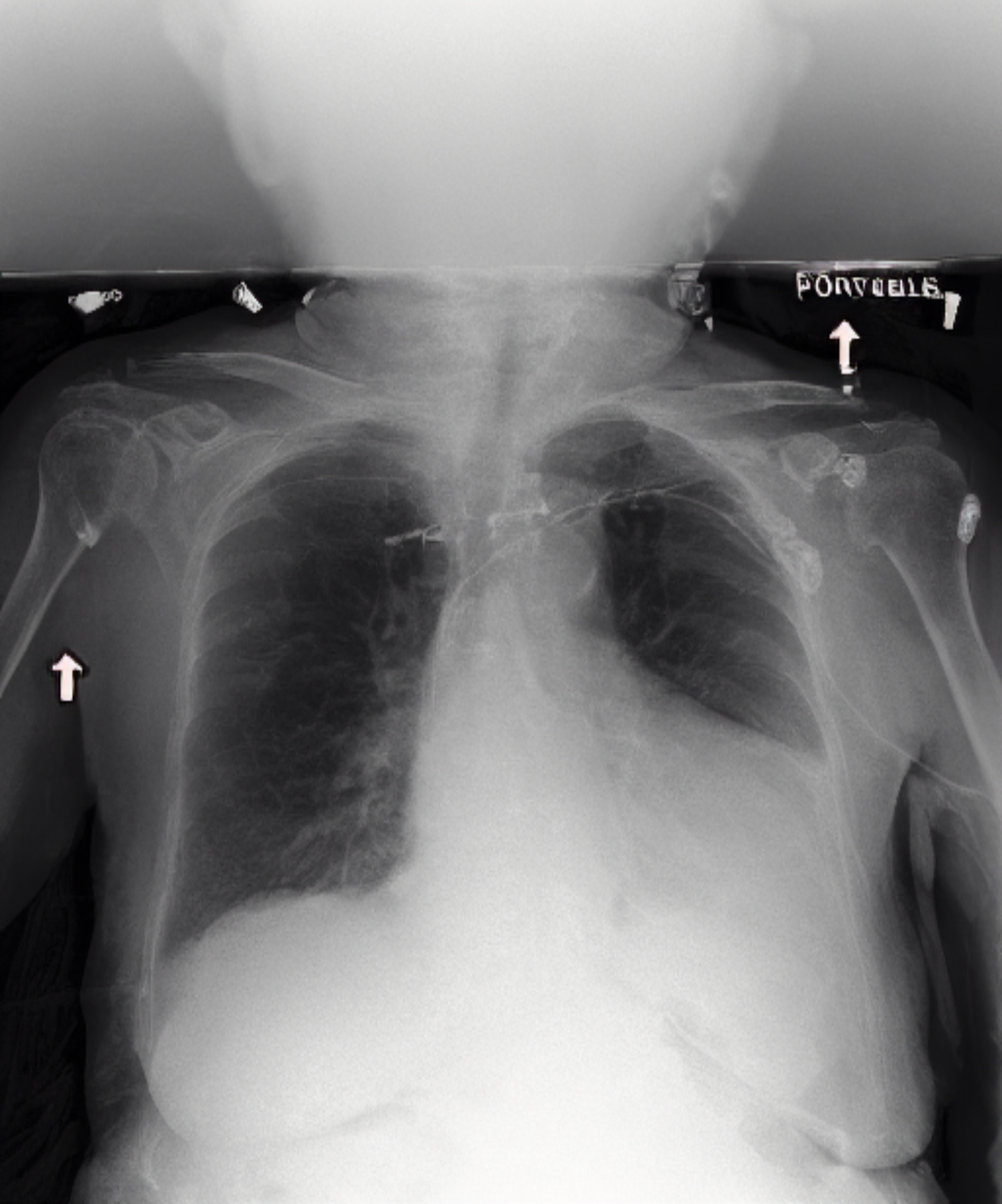} 
    Similarity with the GT: 78.6\%
    &\includegraphics[width=0.2\textwidth]{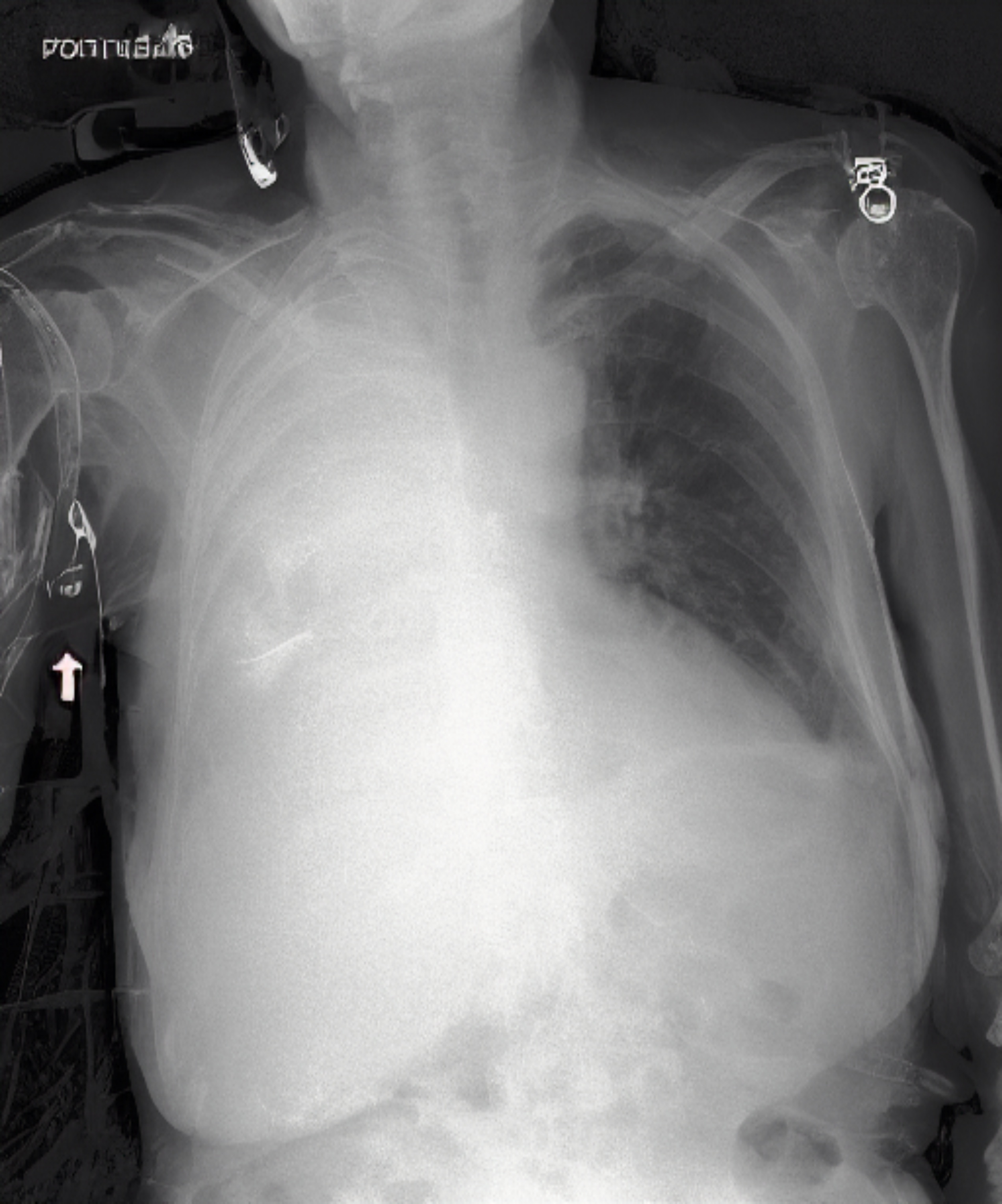}
     Similarity with the GT: 81.9\%
    \\
    \hline
    \multicolumn{4}{|c|}{Generated with A False Report} \\
    \hline
    \includegraphics[width=0.2\textwidth]{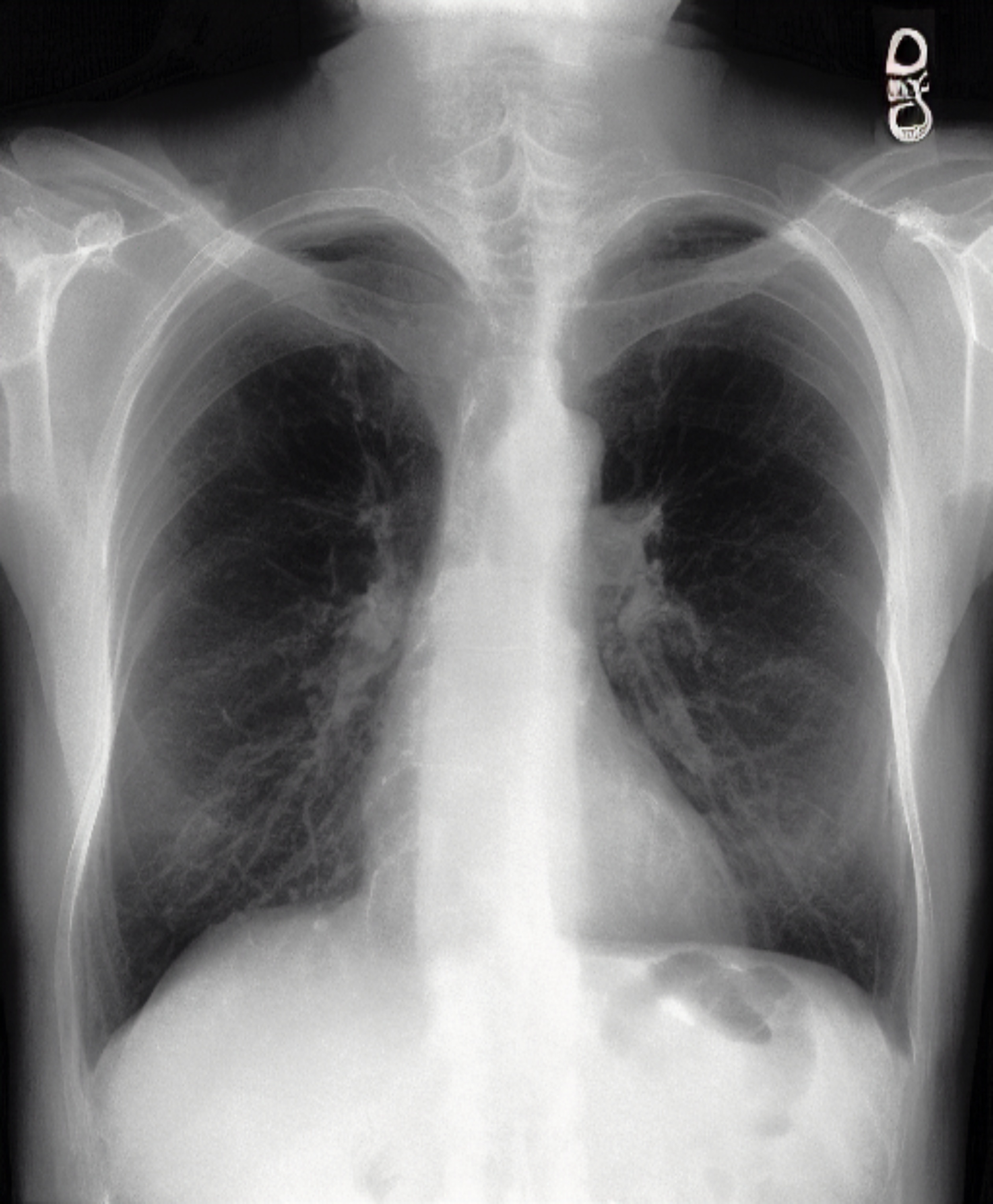} 
     Similarity with the GT: 84.2\%
    & \includegraphics[width=0.2\textwidth]{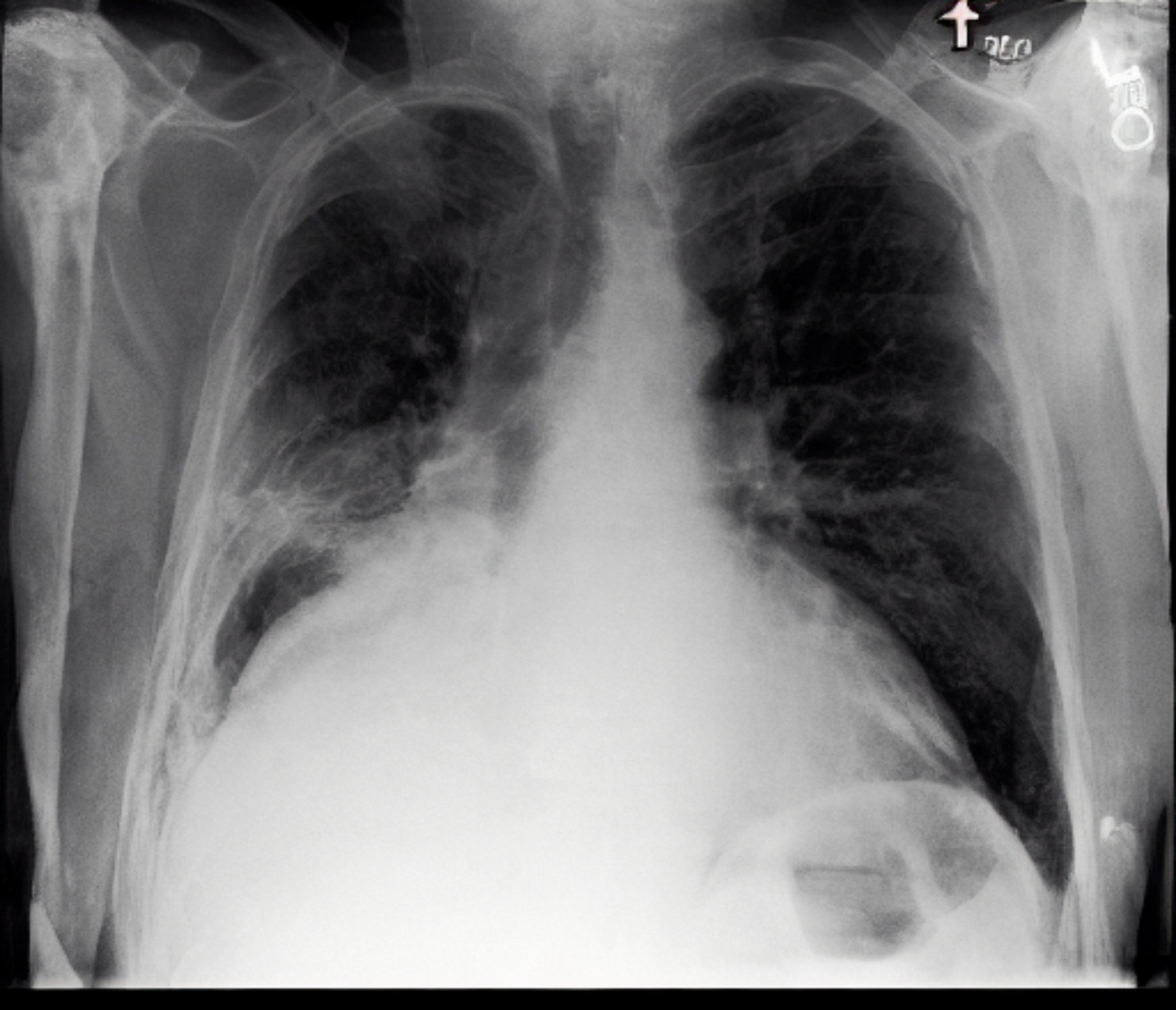} 
     Similarity with the GT: 79.6\%
    & \includegraphics[width=0.2\textwidth]{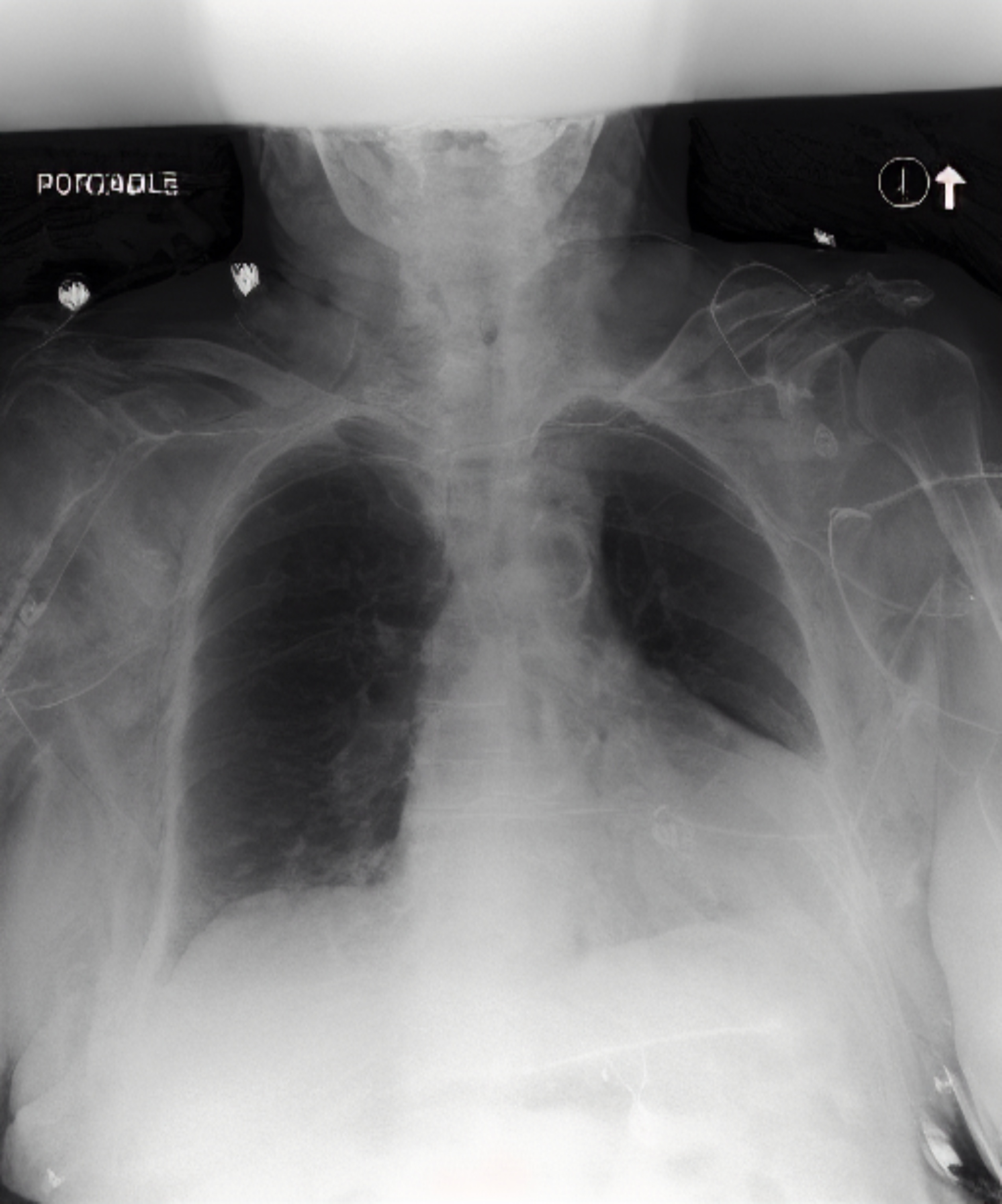} 
     Similarity with the GT: 72.7\%
    & \includegraphics[width=0.2\textwidth]{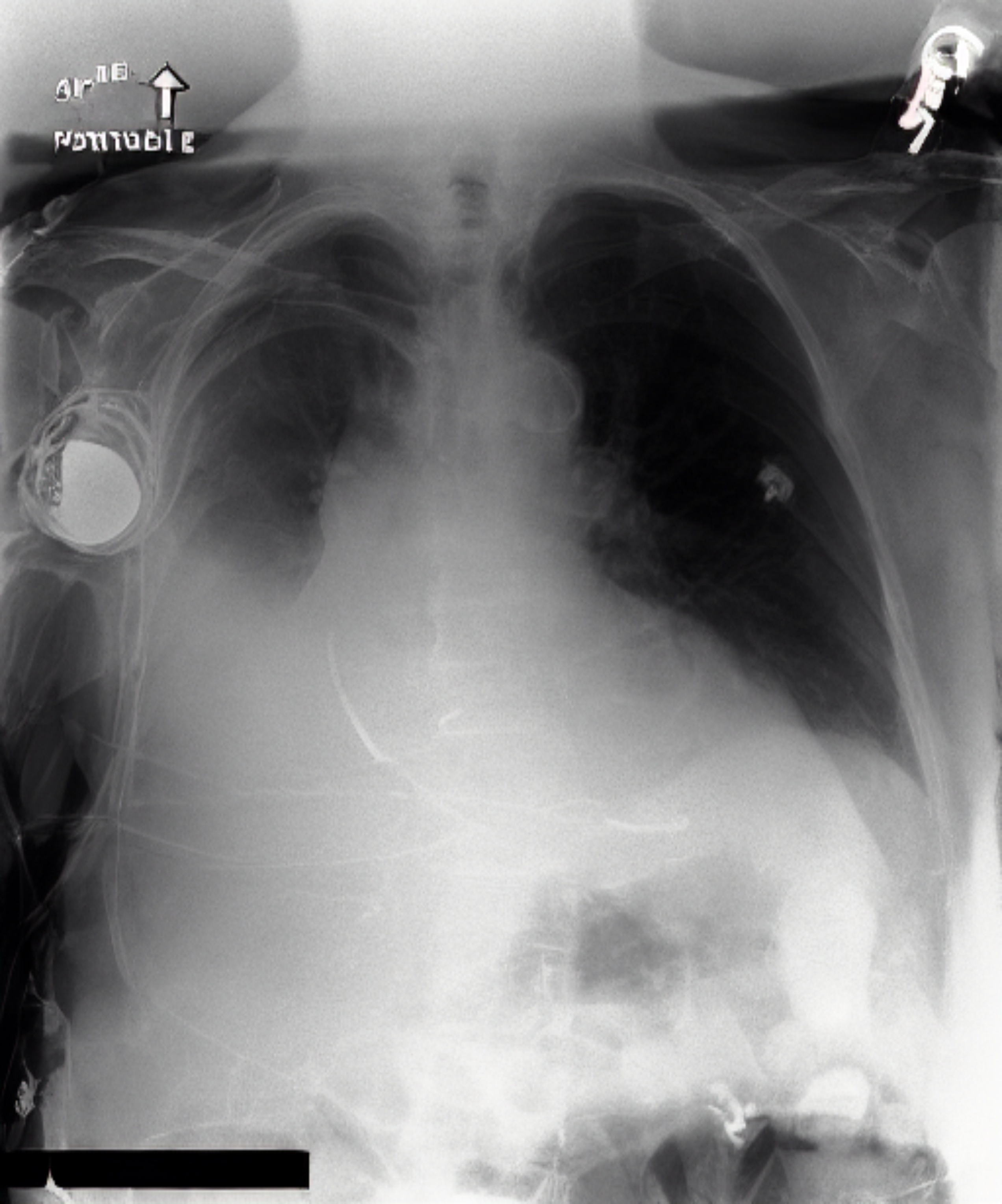} 
     Similarity with the GT: 67.3\%
    \\
    \hline
    \end{tabular}
    \label{tab:result_cn_1}
\end{table}

\begin{table}[]
    \centering
    \caption{The results of the CXR generative model using two different text prompts. The ground truth report is in the middle column, and a random CXR report is in the last column.}
    \begin{tabular}{p{3.5cm}|p{5.2cm}|p{4cm}}
    % \hline
    \centering\includegraphics[width=0.2\textwidth]{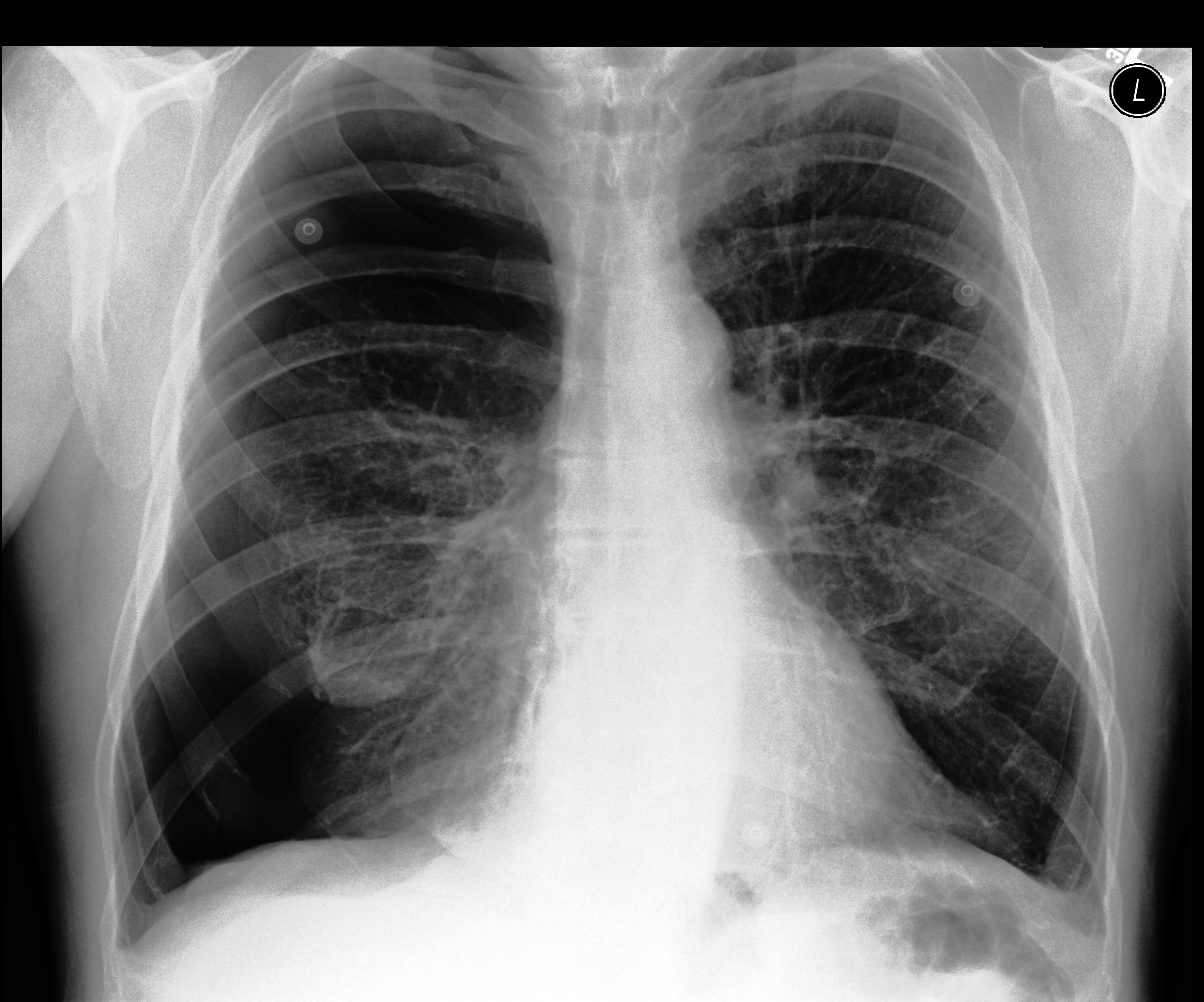}
    
            Original Image
            \label{fig:original}
     &  \centering \includegraphics[width=0.2\textwidth]{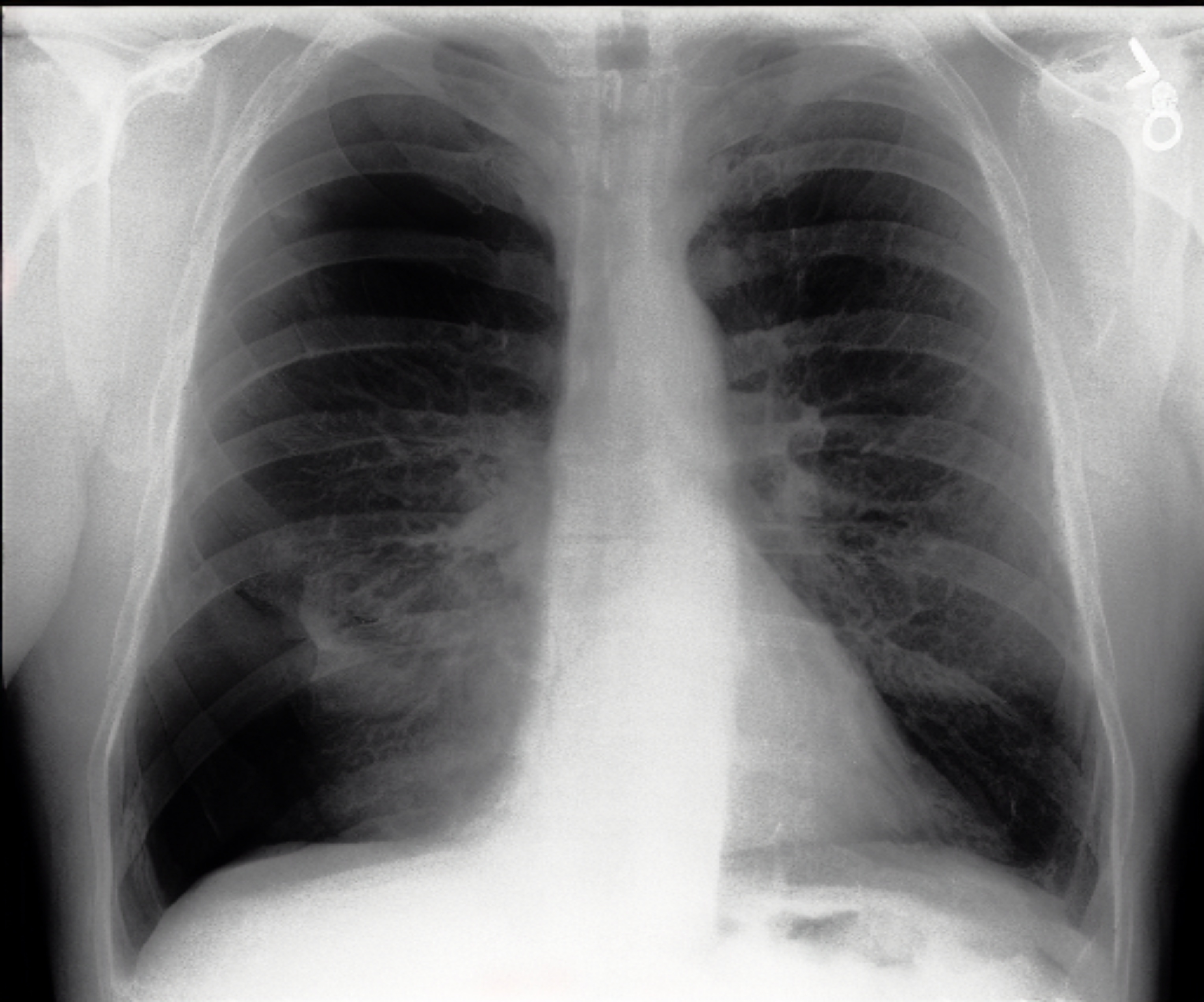}
            % \caption{Caption}
            \label{fig:sampleCN1}
    
     Generated Image by \textit{Ground Truth Report: }Large right-sided pneumothorax with mild leftward shift of mediastinal structures indicative of tension. 
     &
     \centering
     \includegraphics[width=0.2\textwidth]{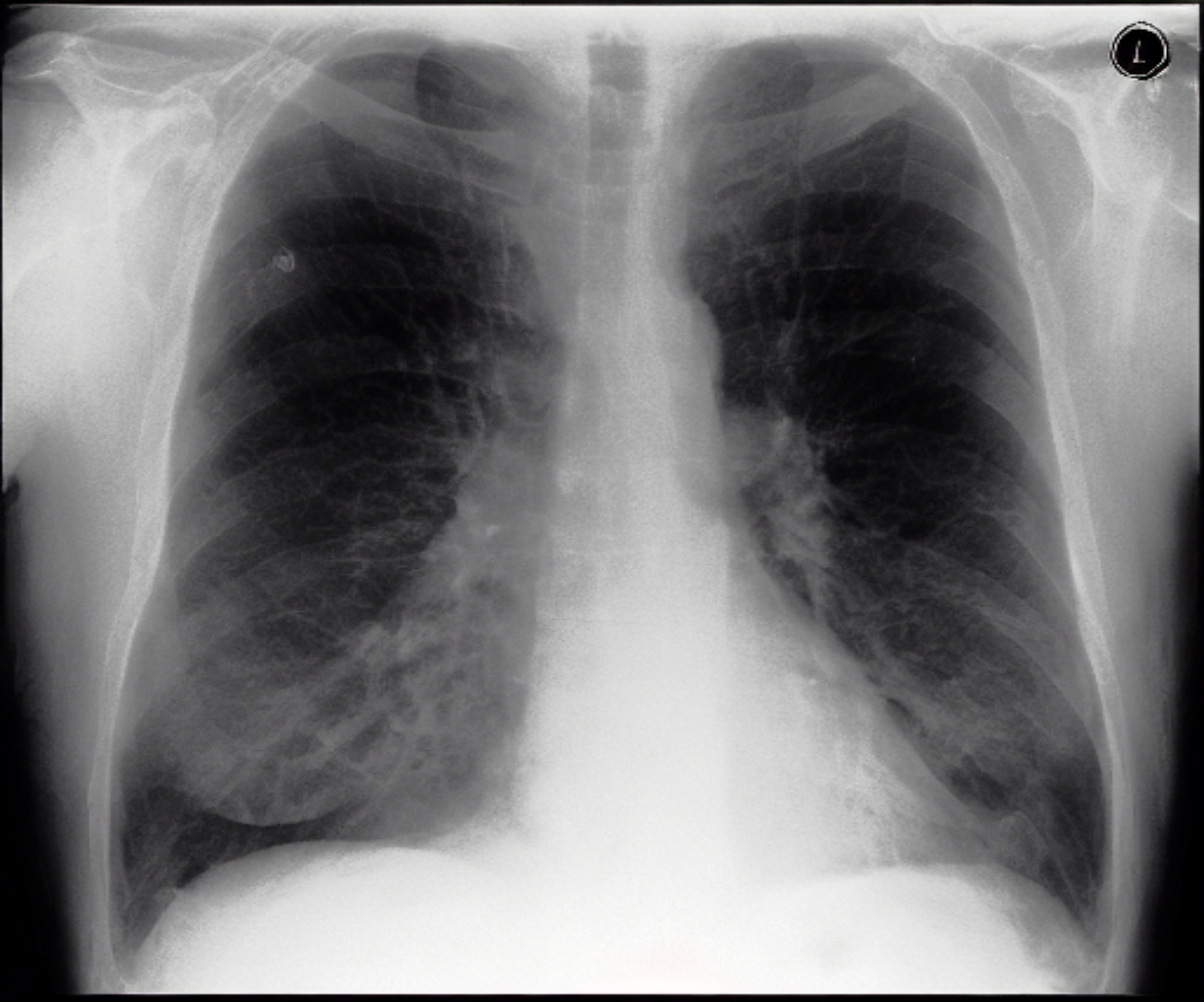}
            % \caption{Caption}
            \label{fig:sampleCN2}
            
     Generated Image by \textit{Random Report: }Cardiomegaly with mild pulmonary vascular congestion. \\
        % \hline
    \end{tabular}
    \label{tab:result_cn_2}
\end{table}

Table \ref{tab:result_cn_1} presents four samples generated using our method. 
In the third row, the generated images result from feeding the binary lung masks and their paired text reports (from the MIMIC-CXR dataset) into the model. The final row illustrates generated images created using the same binary masks but paired with random, unrelated reports.
One challenge observed in these generations is the presence of artifacts around the periphery of the images. While these artifacts are random and irrelevant to the study's focus, they do not affect our main objective, as we concentrate on the lung regions and their associated pathologies. The trained model demonstrates strong capabilities in preserving anatomical structures. For example, in the fourth sample, the right lung in the original image is completely opaque. However, when paired with a random report that does not describe this opacity, the model still synthesizes it accurately, highlighting its robustness in anatomical reconstruction.

For a clearer demonstration of the model's performance with varying reports, Table \ref{tab:result_cn_2} presents an additional example. The middle column displays an image generated using the ground truth report, while the last column shows an image generated using a random, unrelated report. Images generated with the ground truth report closely resemble the original input image, while those generated with incorrect pathology descriptions do not. Notably, when provided with a random report mentioning ``cardiomegaly," the model expanded the heart region in the generated image, reflecting its ability to interpret and map text input to corresponding anatomical features effectively.

Accurately generating features from text enables a robust comparison of feature-level similarity within regions of interest, facilitating the evaluation of the reliability score and validating the alignment between the report and the image.

\section{Training Protocols and Datasets}
\label{sec:training}

\begin{figure}[t!]
    \centering
    % \def\svgwidth{\linewidth}
    % \includesvg[width=\linewidth]{figure_9_Database_flowchart.svg}
    \includegraphics[width=\linewidth]{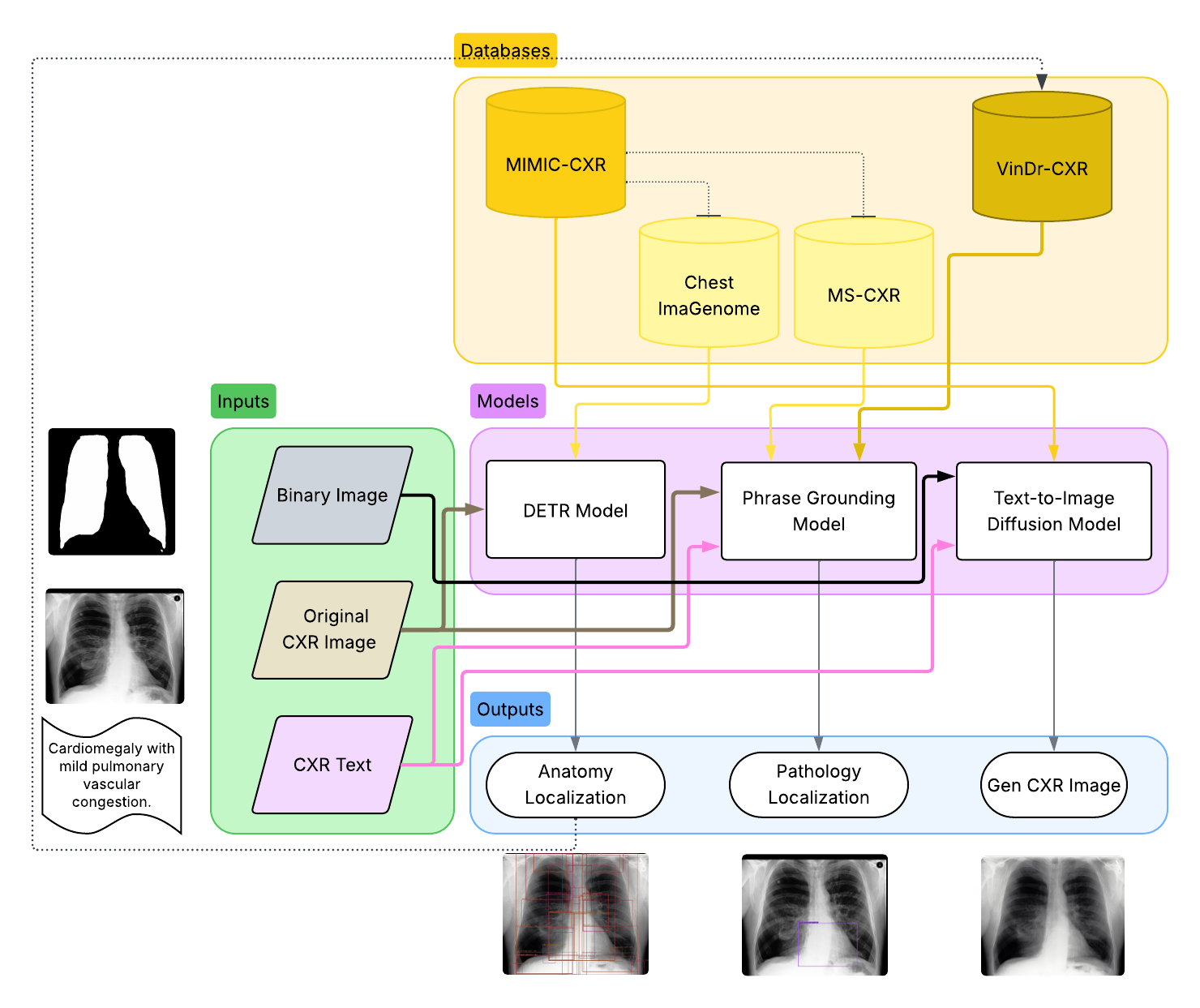}
    \caption{Overview of the training flowchart for the VICCA framework. Each model is trained independently using specific datasets. The phrase grounding model is trained on the MS-CXR and the VinDr-CXR dataset, which is enriched with anatomical region annotations derived using a DETR model trained on the chest ImaGenome dataset. The MIMIC-CXR dataset serves as the source dataset for both MS-CXR and chest ImaGenome, providing the base images, while the latter two add complementary modalities. The generative model is trained on MIMIC-CXR using paired chest X-ray images and corresponding radiology reports.}
    \label{fig:flowchart}
\end{figure}

Each model is trained independently using different datasets. Figure \ref{fig:flowchart} illustrates the overall training pipeline, showing the input and output modalities for each model along with their corresponding training datasets.
Table \ref{tab:training} summarizes the distribution of training and validation datasets for each model, detailing the number of images and their associated annotations. Across all models, only chest X-ray images in the Posterior-Anterior (PA) or Anterior-Posterior (AP) views are included, while those in the lateral view are excluded. Additionally, images labeled as ``No Finding" in the VinDr-CXR dataset are excluded from the training process.

In the case of the MIMIC-CXR dataset, which contains approximately 53,000 images labeled as ``No Finding", we exclude 20,000 of these images. This adjustment is made to balance the dataset more effectively with other pathology labels, ensuring equal representation during training.

\begin{table}[]
    \centering
    \caption{Dataset distribution used for training and validation across the models, including the number of images and their paired annotations.}
    \begin{tabular}{l|ccc}
    \hline
    \multicolumn{4}{c}{Phrase Grounding Model}\\
    \hline
    Dataset & Train & Validation & Test\\
    \hline
    MS-CXR \cite{boecking2022making} & 817 & 169 & 176\\
    VinDr-CXR \cite{VinDrCXRnguyen2022vindrcxropendatasetchest} & 34367 & 3000 & 2697\\
    \hline
    \multicolumn{4}{c}{Stable Diffusion}\\
    \hline
    MIMIC-CXR \cite{32johnson2019mimiccxrjpg} & 207827 & 2991 & 2189\\
    \hline
    \multicolumn{4}{c}{Automatic Lung Segmentation}\\
    \hline
    Chest ImaGenome \cite{79imagenomeWu2021ChestID} & 166521 & 23593& 47393\\
    \hline
    \end{tabular}
    
    \label{tab:training}
\end{table}

\section{Result and Validation}
\label{sec:results}

Developing robust and interpretable evaluation metrics is critical in clinical AI. Prior studies have proposed innovative scoring frameworks to enhance model transparency and interpretability \cite{suggested_cite_3}, an objective we also pursue through our dual-scoring system. 
To ensure meaningful evaluation of the full pipeline, we first assess each component independently. While reaching state-of-the-art performance is desirable, our emphasis is on achieving results that are reasonably aligned with existing benchmarks. This approach reflects our focus on the practical contribution of each model to the overall performance and reliability of the complete pipeline.

% Designing robust and interpretable evaluation metrics is essential in clinical AI. Prior studies have introduced novel scoring frameworks to improve transparency and interpretability in neural models, a goal we similarly pursue through our dual-scoring system \cite{suggested_cite_3}.
% To effectively evaluate the entire pipeline, it is crucial to first assess each component individually. While achieving state-of-the-art performance would be ideal, our primary focus is on achieving results that are reasonably close to existing benchmarks. This is because each model's contribution we made is essential for the overall evaluation and functionality of the complete pipeline.

\subsection{Chest X-ray Phrase Grounding}\label{subsec:PGCXR_Val}

The standard approach for evaluating a phrase grounding model involves mean Average Precision (mAP) for token understanding and mean Intersection over Union (mIoU) for object detection. To this end, we evaluate the model using the test sets of two datasets, with detailed information provided in Table \ref{tab:training}. For the VinDr-CXR dataset, captions are generated based on the pathology and the closest associated anatomical region. The mAP is calculated using the probability scores for each detection, and the results are summarized in Table \ref{tab:result_PG_AP}.

The results for the VinDr-CXR dataset, presented in Table \ref{tab:result_PG_AP}, reflect the performance of benchmark models on the object detection task. Since our method of generating captions based on anatomical regions and pathologies is a novel contribution, direct comparisons with other models for the phrase grounding task are not possible. Nevertheless, our approach outperforms the benchmark models on the test set for object detection in this dataset.

For the MS-CXR dataset, the results pertain to the phrase grounding task. Our model achieves comparable performance to the benchmarks in terms of mAP and surpasses them in mIoU. These findings highlight the effectiveness of our model for CXR phrase grounding tasks, demonstrating its capability to localize and interpret pathologies with high precision.

\begin{table}[]
    \centering
    \caption{Phrase Grounding model performance against VinDR-CXR and MS-CXR datasets.}
    \begin{tabular}{c|cc}
    \hline
    \multicolumn{3}{c}{VinDr-CXR}\\
    \hline
    Model & mAP $\uparrow$ & mIoU $\uparrow$\\
    \hline
    ChEX \cite{muller2024chex} & 14.12 $\pm$ 0.95 & - \\
    BioVIL \cite{boecking2022making} & 2.82 $\pm$ 0.25 & - \\
    VICCA (Ours) & \textbf{34.36 $\pm$ 1.79} & \textbf{38.32 $\pm$ 2.7}\\
    \hline
    \multicolumn{3}{c}{MS-CXR}\\
    \hline
    ChEX \cite{muller2024chex} & \textbf{44.47 $\pm$ 2.21} & 47.52 $\pm$ 1.45 \\
    TransVG \cite{deng2021transvg} & 44.05 $\pm$ 2.63 & 53.51 $\pm$ 1.53\\
    BioVIL \cite{boecking2022making} & 18.62 $\pm$ 1.37 & 28.57 $\pm$ 1.31\\
    VICCA (Ours) & 41.67 $\pm$ 0.69 & \textbf{55.27 $\pm$ 2.36}\\
    \hline
    \end{tabular}
    \label{tab:result_PG_AP}
\end{table}

\subsection{Stable Diffusion for CXR Generation}

To evaluate the generative model, we compare its performance against benchmark models using three metrics: Multi-Scale Structural Similarity Index (MS-SSIM), Dice score, and Frechet Inception Distance (FID). These metrics collectively assess the quality and anatomical fidelity of the generated images.

For evaluation, we generated 10 samples for each of the 2,065 reports in the test set, resulting in a total of 20,650 generated images.
For the FID evaluation, we use all generated samples. For the MS-SSIM and Dice evaluations, we employ an automated selection approach to identify the most representative sample among the 10 generated images. The selected sample is the one with the highest IoU score and the closest pathology classification match to the original CXR image.
The results, presented in Table \ref{tab:fidscore}, indicate that our model achieved the second-best FID score while outperforming all benchmarks in MS-SSIM and Dice. These findings highlight the superior quality of our model's generated images, particularly in preserving anatomical structures and ensuring alignment with pathology information.

The results highlight the effectiveness of our inpainting technique during training, demonstrating that a model originally trained on natural images can be successfully adapted for CXR images by utilizing binary lung segmentation to maintain anatomical consistency and a specialized text encoder for medical contexts. In comparison, the Cheff model with the best FID score generates images solely based on text input and lacks this anatomical precision. Furthermore, our model surpasses the performance of the XReal model, which also focuses on preserving anatomical accuracy, further validating our approach.

\begin{table}[]
    \centering
    \caption{The comparison of FID scores among three models.}
    \begin{tabular}{l|ccc}
    \hline
    \textbf{Model}& 
         MS-SSIM $\uparrow$ & Dice $\uparrow$ & FID $\downarrow$\\
         \hline
         Cheff \cite{36cheffweber2023cascaded} & 0.415 & 0.5 &\textbf{24.64}\\
         RoentGen \cite{6chambon2022roentgen} & 0.386 & 0.631 & 82.14\\
         XReal \cite{hashmi2024xreal}& 0.701 & 0.838 & 55.12\\
         VICCA (Ours)& \textbf{0.71}&\textbf{0.841} & 35.76\\
         \hline
    \end{tabular}
    \label{tab:fidscore}
\end{table}

While the quantitative metrics used are valuable for validating the generative model's performance, the sensitivity of the medical domain necessitates a deeper focus on medical validation. In this study, we design a validation framework that does not rely on direct radiologist feedback, aiming to develop an expert-independent assessment pipeline. To ensure the model's medical reliability, we leverage well-annotated datasets and standardized evaluation metrics, establishing a structured and objective validation process that enhances reproducibility and clinical applicability.

Before evaluating the entire pipeline, it is crucial to validate the generative model specifically through medical criteria, as it serves as a reference for assessing localization accuracy. To ensure the robustness of the generated outputs, we add a novel system to the framework. This system is designed to evaluate the anatomical and pathological aspects of the model. This additional validation ensures that the generated images align with the medical standards required for accurate interpretation and assessment.

\subsection{Validation Process}

The validation process for the generative model in this research is inspired by the way experts acquire knowledge. It centers on evaluating the fundamental aspects of anatomical regions and correlating the anomalous parts of the image with the corresponding pathology. Consequently, our first step is to establish a method for validating the anatomical structures of the generated images. This ensures that the generated outputs maintain consistency with real CXR anatomical accuracy.

\subsubsection{Anatomical Validation}

A typical chest X-ray (CXR) image consists of approximately 36 distinct anatomical regions, such as the ``right upper lung zone," ``right mid lung zone," ``right atrium," ``descending aorta," ``carina," ``left upper abdomen," and others. Identifying these anatomical regions is a straightforward task, thanks to the availability of the Chest ImaGenome dataset \cite{14physionetGoldberger2000-si, 79imagenomeWu2021ChestID}, which provides radiologist-annotated bounding boxes for these 36 regions within the MIMIC-CXR dataset. Leveraging this dataset, we successfully train a robust region detector using the DETR model \cite{detr}.
A notable advantage of the DETR model is its capacity to detect all class objects simultaneously, which is particularly useful given the close proximity of anatomical features in CXR images. The model demonstrates rapid convergence, achieving a DETR-like loss of 3.5030 and a generalized \textbf{IoU loss} of 0.29 after just 14 epochs. On the test set, the IoU accuracy reaches an impressive 76\%.

\begin{figure}
    \centering
    \includegraphics[width=0.5\linewidth]{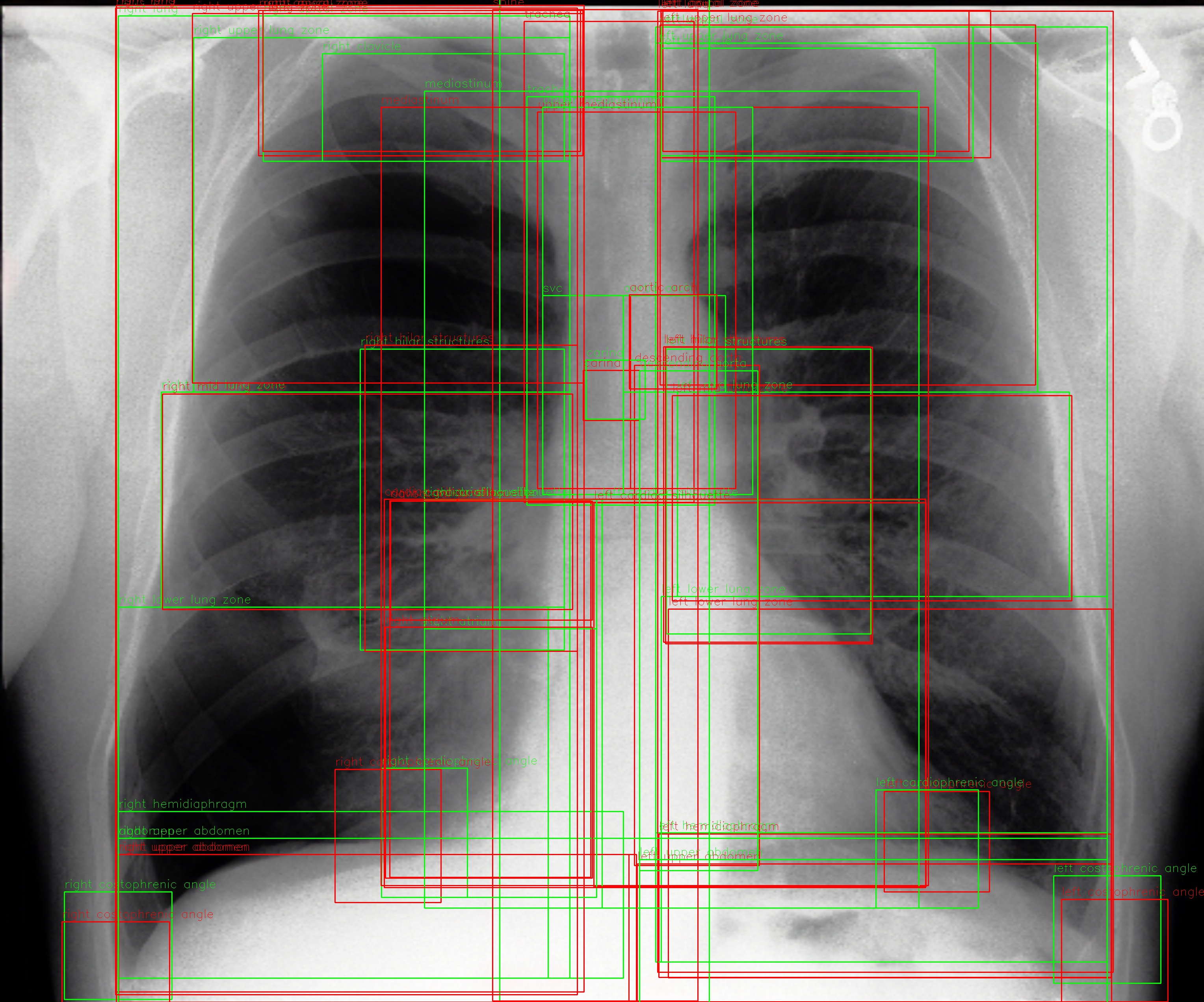}
    \caption{Ground truth anatomical regions in green and automatical detection of the anatomical regions on generated CXR in red. The mIoU of the bounding boxes is 76\%.}
    \label{fig:result_detr}
\end{figure}

Subsequently, we apply the trained DETR model to the generated CXRs to obtain bounding boxes and evaluate their IoU against the ground truth. The IoU score for the generated CXR set is 68\%, validating the generative model's effectiveness in preserving anatomical structures.
Figure \ref{fig:result_detr} demonstrates the performance of the trained DETR model on a generated CXR image. The red bounding boxes represent the automatically detected anatomical regions, while the green bounding boxes depict the ground truth annotations from the Chest ImaGenome dataset.
Since generated images often contain slight spatial shifts compared to ground truth images, we address this issue by aligning the bounding boxes based on the superior vena cava (SVC) region, which is typically central in CXR images. By detecting the positional shift of the SVC in both images, we adjust the bounding boxes accordingly.
The mIoU score between the detected and ground truth bounding boxes is 76.2\%, indicating a high degree of overlap and alignment for the majority of anatomical regions.

\subsubsection{Pathological Validation}

To validate the pathological accuracy of the generated CXR images, we utilize the TorchXrayVision library for pathology classification \cite{Cohen2022xrvtorchxrayvision}.
While various datasets provide annotations for 14 to 15 thoracic diseases, many existing classification models are either outdated or limited to a subset of these classes. TorchXrayVision overcomes these limitations by leveraging multiple datasets for comprehensive pathology classification. Using this library, we classify the pathologies in the generated CXRs and compare the detected pathologies' accuracy with those in the original images. The generative model achieves an average classification accuracy of 88.53\% on the generated test set.

These results highlight our generative model's capability to produce CXR images that accurately reflect pathological features while maintaining anatomical fidelity.

\subsection{VICCA Model Result}

\begin{figure}[t!]
    \centering
    \includegraphics[width=\linewidth]{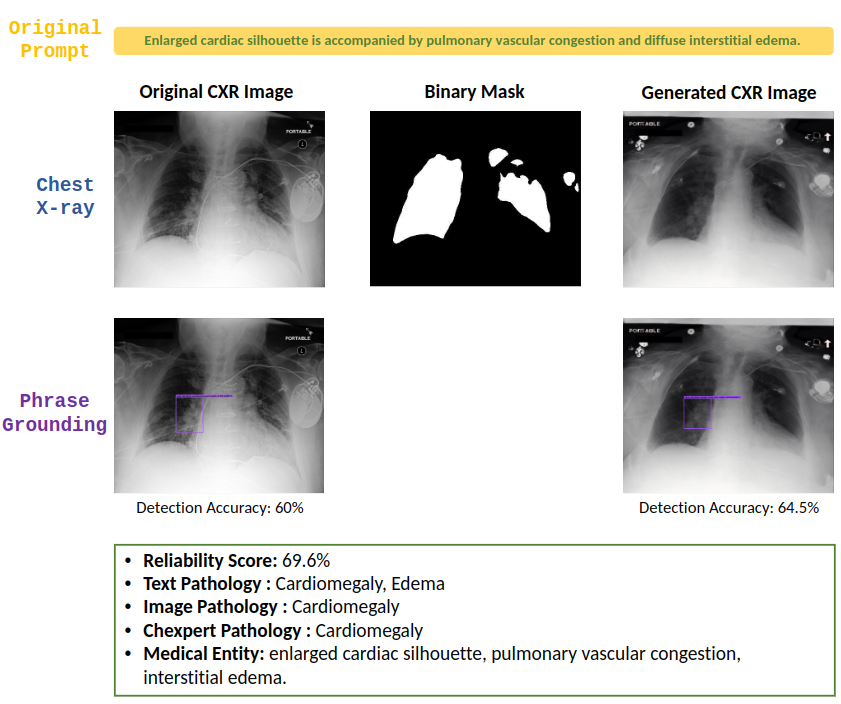}
    \caption{An example output of our VICCA model.}
    \label{fig:pp_res}    
\end{figure}

Figure \ref{fig:pp_res} presents an example result from our VICCA model. Alongside the two scoring metrics (detection accuracy and reliability score) the pipeline also provides additional spatial information regarding the pathology extracted from the input text using our previous model \cite{MCSE}, which identifies relevant medical entities as well. The image pathology is classified using the TorchXRayVision model \cite{Cohen2022xrvtorchxrayvision}, which categorizes CXR images into 14 pathology classes. Since both the original image and the prompt are derived from the MIMIC-CXR dataset, we also include CheXpert pathology annotations \cite{irvin2019chexpertlargechestradiograph} from the official dataset for comparison.

\subsection{Pipeline Evaluation}\label{subsec:pipline_eval}

With the generative model validated and the phrase grounding model demonstrating good precision based on quantitative results, we can now address the primary goal of this research, providing a reliability score for the automatically localized regions.

To achieve this, we translate the bounding boxes from the phrase grounding model onto the generated images, ensuring the exact same regions are evaluated. An encoder trained specifically on chest X-ray images is then used to extract features from these bounding box regions. 
% \textcolor{teal}{\textbf{Comment:} Meanwhile, there's a typo in the last line of page 22 — 'twofold' appears to be incorrect.}
To evaluate the similarity between the original and generated regions, we use a two-fold approach. First, we analyze the pixel-level features by calculating the MS-SSIM. Then, we use an image encoder trained specifically on CXR images to extract high-level features, and we compute the $\chi^2$ value as a quantitative measure of correspondence.

Since no direct reference exists for this evaluation, we design a two-step methodology to interpret both the MS-SSIM and $\chi^2$ metrics effectively:

\begin{enumerate}
    \item \textbf{Using the Original Report:} 
    \begin{itemize}
        \item Bounding boxes are detected based on the original report, and both MS-SSIM and $\chi^2$ values are calculated by comparing the corresponding regions in the original and generated images.
        \item A $\chi^2$ value closer to zero indicates a high feature similarity, while an MS-SSIM value closer to one signifies a strong structural resemblance between the regions.
    \end{itemize}
    \item \textbf{Using a Random Report:} 
    \begin{itemize}
        \item The process is repeated using a random report unrelated to the original image.
        \item The expectation is that the MS-SSIM values would be significantly lower, and the $\chi^2$ values significantly higher, confirming a reduced similarity between the localized and generated regions when the report does not correspond to the actual content.
    \end{itemize}
\end{enumerate}

\begin{table}[]
    \centering
    \begin{tabular}{cc}
        % \resizebox{0.4\textwidth}{!}{\input{ms_ssim_boxplot.pgf}} &  \resizebox{0.55\textwidth}{!}{\input{chi_square_hist.pgf}}\\
        \includegraphics[trim={0.6cm 0 0 0}, clip, width=0.37\linewidth]{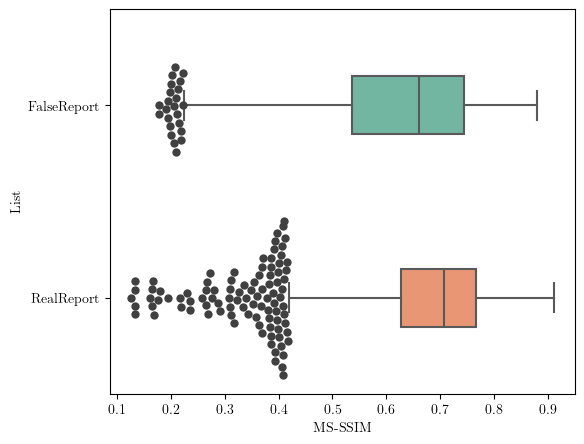}
        &
        \includegraphics[width=0.5\linewidth]{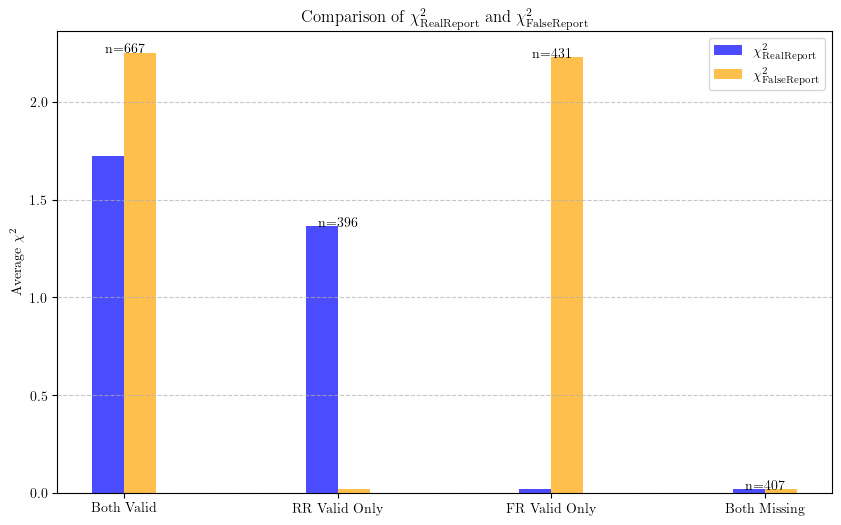}\\
    \end{tabular}
    \captionof{figure}{
The similarity comparison between the localization and the generated CXR images is evaluated using MS-SSIM and $\chi^2$ calculations based on two text inputs: the Real Report (RR), which represents the ground truth associated with the CXR image, and a False Report (FR), which has no correlation with the RR.
The $\chi^2$ plot on the right categorizes the results into four scenarios based on the output of the phrase grounding model: Both RR and FR produce valid localizations, only RR produces a valid localization, only FR produces a valid localization, neither RR nor FR yields a valid localization.
The MS-SSIM boxplot on the left illustrates the distribution of similarity scores for each scenario, showing the density of evaluation for both the real and false reports. }
    \label{fig:ssim_chi}
\end{table}

Figure \ref{fig:ssim_chi} illustrates the results of this comparison based on the average MS-SSIM and the $\chi^2$ values. 

For MS-SSIM, the similarity scores based on the Real Report indicate a narrower range, with outliers remaining close to the margins, indicating a consistent and high degree of similarity. In contrast, the scores based on the False Report show a wider distribution, with a noticeably lower average similarity, reflecting the reduced alignment between the generated and original images in this scenario. 
The higher similarity scores observed can be attributed to the method used for selecting random reports, which is based on pathology annotations from the MIMIC-CXR dataset. Random reports are considered as acceptable if they lack any explicit similarity to the original reports. However, in certain cases, overlapping visual traits between pathologies, for instance, ``lung opacity" and ``nodule", led to unintentional correlations. These shared visual characteristics in the selected random reports contribute to the detection of higher similarity scores.

Regarding feature similarity using $\chi^2$, analysis across 1901 samples shows the following observations:

\begin{itemize}
    \item 407 samples have low localization probabilities for both the original and random reports, rendering the results inconclusive.
    \item 396 samples show a valid localization only when using the original report.
    \item 431 samples produce more acceptable localization results with the random report compared to the original report.
    \item 667 samples demonstrate significantly lower $\chi^2$ values for the original report compared to the random report, indicating better feature correspondence.
\end{itemize}

Although there are 431 cases where the localization model fails to detect regions accurately based on the original report, the pipeline's performance is evident. Even in these cases, the average $\chi^2$ value for regions localized using the random report is relatively high, reflecting the system's ability to identify and highlight dissimilarities between false report segments in generated and original images. 
This underscores the robustness and sensitivity of the pipeline, even when faced with similarities in the pathology.

\begin{figure}
    \centering
    \includegraphics[width=\linewidth]{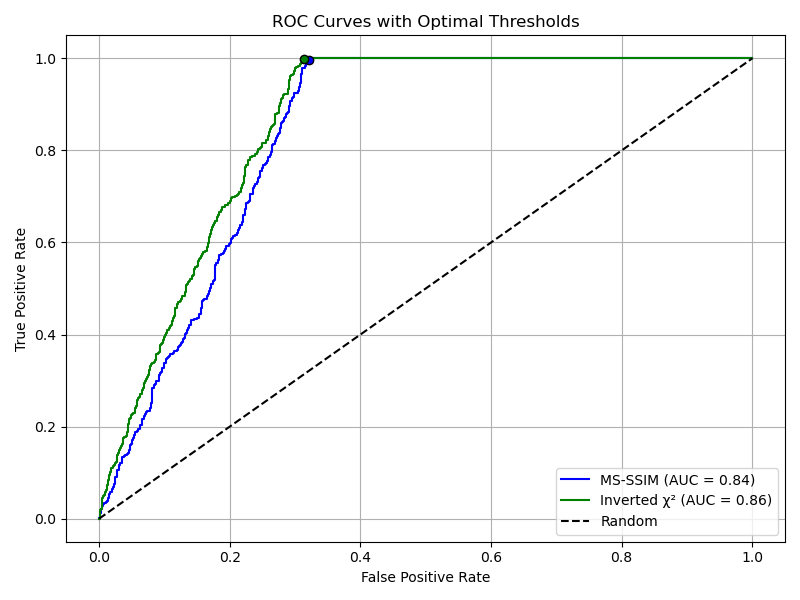}
    \caption{ROC curves for the dual-scoring interpretability system using MS-SSIM and inverted $\chi^2$ metrics. }
    \label{fig:ROC}
\end{figure}

To support the system's utility in clinical decision-making, Figure~\ref{fig:ROC} presents the ROC curves for the MS-SSIM and $\chi^2$ scoring metrics. As explicit interpretability labels are unavailable, we assign a ground truth label of 1 to cases where the phrase grounding model detects regions in both the original and synthesized images, and 0 otherwise. To determine optimal thresholds, we used Youden's J statistic \cite{Youden1950-hl} (J = True Positive Rate (TPR)  – False Positive Rate (FPR)), which identifies the point on the ROC curve that maximizes the difference between the true positive rate and the false positive rate. Based on this criterion, the optimal thresholds were found to be 0.299 for MS-SSIM and 0.123 for $\chi^2$, with both metrics achieving a true positive rate of 1.00 and a false positive rate of approximately 0.31. This means that while all interpretable cases are correctly identified, about one-third of non-interpretable cases are misclassified.

These results suggest that the dual-scoring system prioritizes sensitivity, reliably capturing all interpretable cases, even at the cost of moderate over-inclusiveness. In clinical contexts, this trade-off is acceptable and even preferable, as ensuring that no interpretable region is missed is often more critical than avoiding false positives.

\begin{figure}
    \centering
    \includegraphics[width=\linewidth]{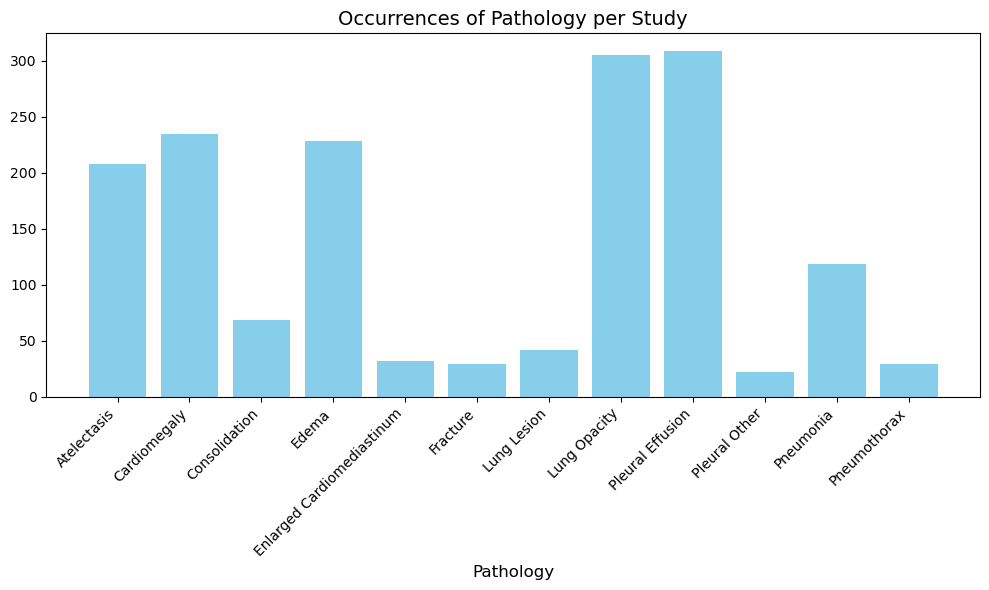}
    % \resizebox{0.49\textwidth}{!}{\input{Occurrences.pgf}}
    \caption{Occurrences of pathologies in the test dataset, including 2018 studies. The most frequent occurrences are observed for ``Lung Opacity" and ``Pleural Effusion," followed by ``Cardiomegaly" and ``Edema."}
    \label{fig:occur}
\end{figure}

Finally, we present the results for each pathology class. Figure \ref{fig:occur} illustrates the frequency of each pathology in our test dataset. The most frequent occurrences are observed for ``Lung Opacity" and ``Pleural Effusion," followed by ``Cardiomegaly" and ``Edema."

\begin{figure}
    \centering
    % \resizebox{0.49\textwidth}{!}{\input{violin_ssim.pgf}}
    \includegraphics[width=\linewidth]{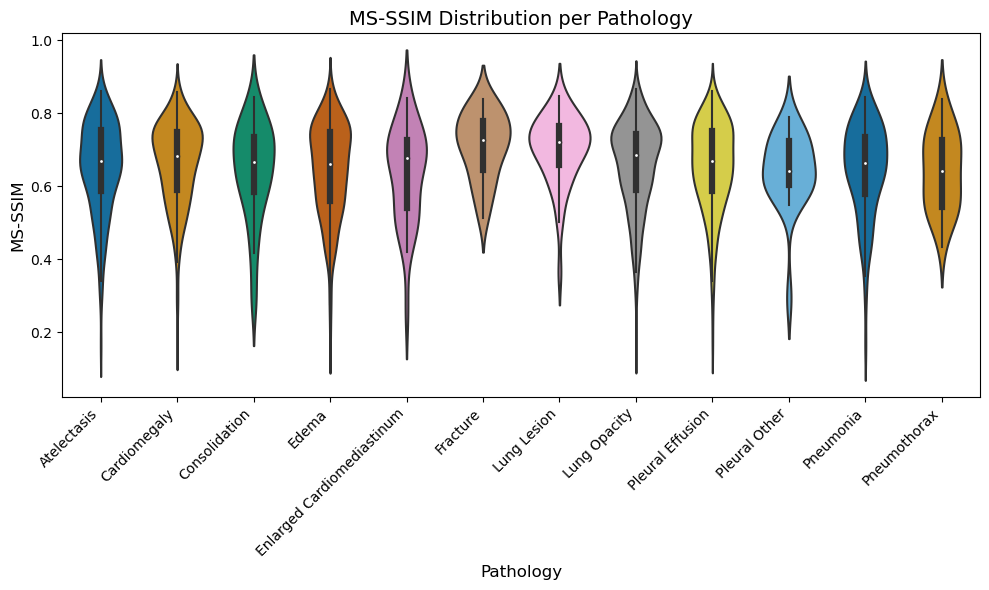}
    \caption{The MS-SSIM distribution per pathology in the test dataset.}
    \label{fig:violin}
\end{figure}

For each pathology, we present the MS-SSIM score for each subject in Figure \ref{fig:violin}. The results show that the highest average scores are associated with ``Lung Lesion" and ``Fracture". Although these pathologies are less frequent in our test set, they are easier to interpret in a CXR image. Based on our findings, the most challenging pathologies to interpret in a CXR image are ``Pneumonia", ``Pneumothorax", and ``Pleural others". Among these, ``Pneumonia" occurs more frequently but remains the most difficult to assess. Additionally, our results indicate that ``Edema" is the most ambiguous class, as it shows the highest variability. This variability could be attributed to the subtle and often similarity in nature of the symptoms in CXR images, which may lead to difficulties in distinguishing them from each other. 
Our overall findings suggest that this interpretive study is crucial for enhancing the understanding and reliability of AI-generated reports. A few case studies are presented in \ref{sec:case_study} to further demonstrate the effectiveness of our VICCA model.

\section{Generalizing VICCA Beyond Chest X-rays}
\label{sec:future_generalization}

While our work focuses on chest X-rays, the VICCA framework is not inherently limited to this modality. Its modular architecture enables potential adaptation to other medical domains, such as brain MRI for acute stroke assessment. For instance, a recently published dataset \cite{MRI} provides brain MRI scans of stroke patients, including lesion volume data and detailed clinical annotations.

By fine-tuning VICCA on such MRI datasets, the phrase grounding module could be trained using lesion segmentation maps aligned with corresponding clinical reports to enable visual interpretation of findings. Simultaneously, the text-to-image diffusion module could be adapted to synthesize pathology-aware MRI images, serving as a visual proxy for report validation. This would, however, require a dedicated text encoder trained on MRI-specific clinical language, as well as an image encoder suited for the MRI modality. Additionally, VICCA's diffusion component would need a pre-processing step to generate binary brain masks to preserve anatomical structure during image synthesis.

Together, these adaptations could enable VICCA to offer dual-score validation, quantifying both localization and semantic consistency, thereby advancing the transparency and interpretability of AI-generated reports in neuroimaging contexts \cite{lei2024autorgbraingroundedreportgenerationMRI}. Nonetheless, these extensions remain hypothetical and warrant further empirical validation.

\section{Conclusion}
\label{sec:discussion}
Recent advancements in chest X-ray radiology studies have prioritized enhancing interpretability, with a significant reliance on large language models for automated report generation. While these models show promising performance, occasional inaccuracies in their predictions limit their clinical applicability. This research introduced a novel approach to achieving a high degree of interpretability by leveraging AI-driven metrics and perspectives that mimic radiologist evaluations without the need for human feedback.

Our framework demonstrates that combining multiple AI models can enhance interpretability by offering visual explanations with confidence scores for localization and introducing interpretative metrics for ROI through generative models. This approach not only supports radiologists by offering AI-assisted for generated reports but also aids general practitioners in areas with limited access to expert radiologists, improving decision-making and patient care, and ultimately improving patient outcomes on a broader scale. Additionally, our model serves as a valuable tool for researchers aiming to improve their report generation models.

In this project, we introduced not just a multimodal pipeline of different AI models, but our work represents a significant step forward in the application of AI to chest X-ray radiology by introducing a concept of collaborative AI decision-making and enhancing the interpretability of their outputs. Our proposed training strategy integrates multiple tasks and datasets, presenting a viable approach to achieve this goal. By leveraging diverse AI models and their respective strengths, we created a robust framework that can provide comprehensive insights and support across various clinical scenarios.

\section{Future Work}
While VICCA demonstrates promising results in assessing the interpretability of AI-generated chest X-ray reports, a limitation of this study is the absence of a formal evaluation by medical experts. Although we designed the framework to function without clinician feedback, future work would involve a comprehensive user study or direct comparison with radiologist assessments to further validate its clinical relevance. Establishing such collaborations remains a challenge due to the complexity of access to medical personnel, but we recognize the importance of involving domain experts to benchmark the system's practical utility.

\section{Code Repository}

Results can be reproduced using the code available in the GitHub repository \url{https://github.com/sayeh1994/vicca.git}.  The code will be accessible after submission.

\section{Acknowledgment}
We would like to acknowledge that all the computations presented in this paper are performed using the infrastructure (\url{https://gricad.univ-grenoble-alpes.fr}), which is supported by Grenoble research communities.

\section{Declaration of generative AI and AI-assisted technologies in the writing process}

During the preparation of this work, the author used the ChatGPT service in order to improve the readability and language of the work. After using this tool/service, the author reviewed and edited the content as needed and takes full responsibility for the content of the publication.

\bibliographystyle{unsrtnat}
\bibliography{ref}

% \newpage
% \onecolumn
\appendix

\renewcommand{\thefigure}{A.\arabic{figure}}
\renewcommand{\thetable}{A.\arabic{table}}
\setcounter{figure}{0}
\setcounter{table}{0}

\section{Case Study} \label{sec:case_study}

We present various scenarios and examples encountered during our study to illustrate the application and performance of our project. These include both successful cases and failed cases, which are included to provide deeper insights and facilitate a better understanding of the limitations and challenges of the approach.

\subsection{Case 1: Multiple Pathologies}

\begin{figure}[t!]
    \centering
    \includegraphics[width=\linewidth]{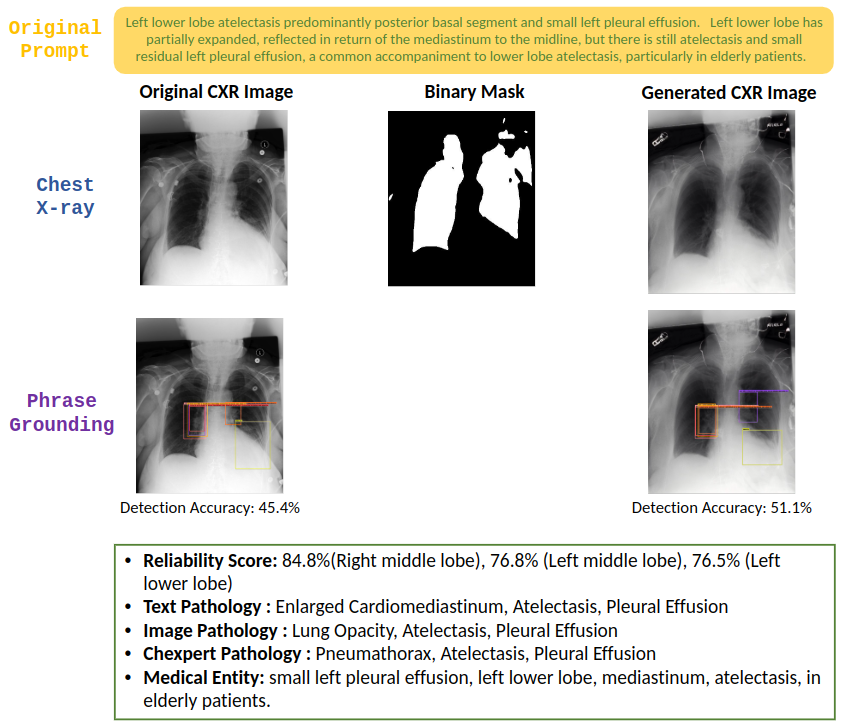}
    \caption{Case 1: Multiple Pathology}
    \label{fig:case_1}    
\end{figure}

In the first scenario presented in figure \ref{fig:case_1}, we examine a case involving multiple pathologies in both the text prompt and the image. Using our phrase grounding visualization, we identify three regions of interest (ROIs) in both the original and generated images. Although the detection probabilities are relatively low, the high similarity between the corresponding ROIs across both images indicates that the prompt can be considered trustworthy.
Additionally, the pathologies identified through the prompt and the image using an AI model closely align with the Chexpert annotations derived from the MIMIC-CXR dataset \cite{30mimic-cxrJohnson2019}. 

\subsection{Case 2: Single Pathology A Success Case}

\begin{figure}[t!]
    \centering
    \includegraphics[width=\linewidth]{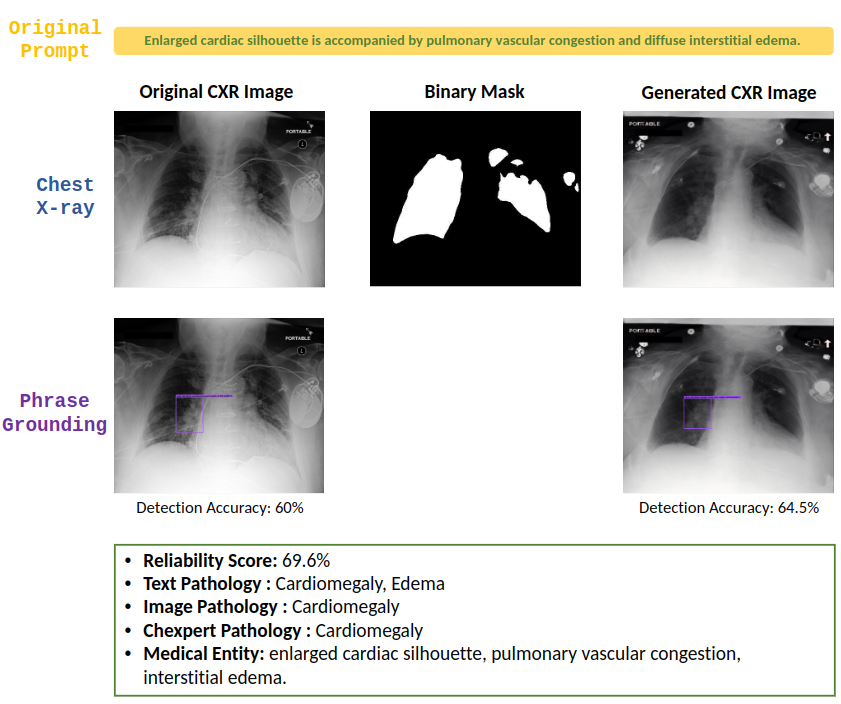}
    \caption{Case 2: Single Pathology A Success Case}
    \label{fig:case_2}    
\end{figure}

In the second case, presented in Figure \ref{fig:case_2}, we analyze a scenario involving a single pathology. While two pathologies are initially identified in the text prompt, both our image-based pathology detection and the Chexpert annotation confirm the presence of only one pathology, maintaining consistency across methods. Moreover, our phrase grounding accurately detects a single pathology, which exhibits a high similarity with the pathology described in the prompt through the generated image. This alignment demonstrates that the prompt is consistent with the image and can be considered trustworthy.

\subsection{Case 3: Single Pathology A Failure Case}

\begin{figure}[t!]
    \centering
    \includegraphics[width=\linewidth]{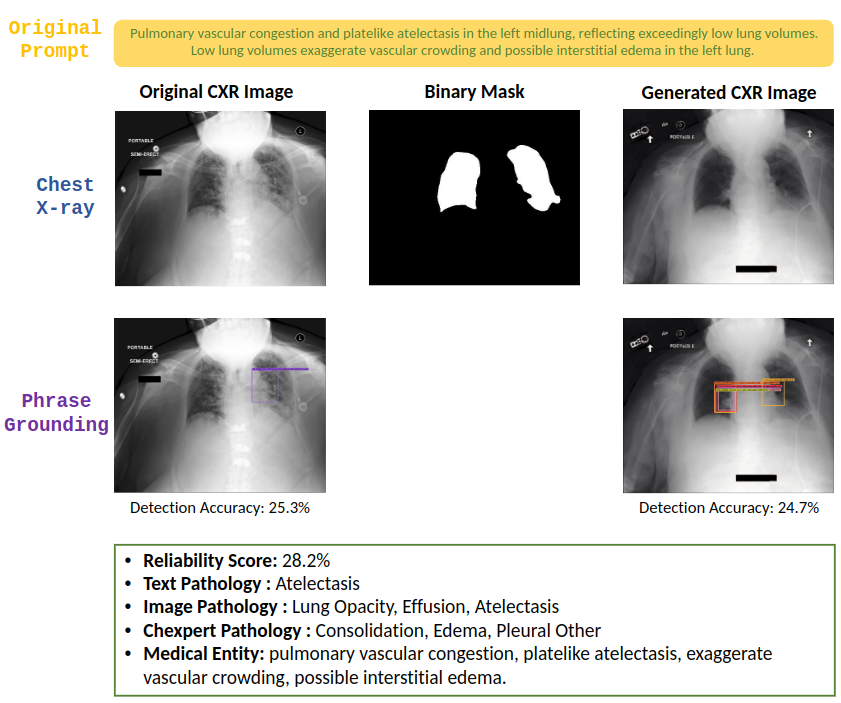}
    \caption{Case 3: Single Pathology A Failure Case}
    \label{fig:case_3}    
\end{figure}

In contrast to the previous case, Figure \ref{fig:case_3} illustrates a failure to correctly align the original prompt with the corresponding image. Phrase grounding on the original CXR image detects only one ROI associated with pathology, whereas pathology detection identifies three distinct pathologies in the image. Similarly, the generated image from the same prompt reveals two ROIs, but their similarity scores with the original image are notably low.

While the AI model identifies only one pathology from the text, this pathology corresponds to one of the detected pathologies in the image. However, these results do not align with the annotated pathologies in the dataset. Despite the prompt being originally written by a radiologist, the system fails to reliably associate it with the correct CXR image. This case highlights challenges in ensuring consistency and accuracy across modalities when relying solely on automated approaches without any interpretation.

\subsection{Case 4: A Success Case of Using False Report}

\begin{figure}[t!]
    \centering
    \includegraphics[width=\linewidth]{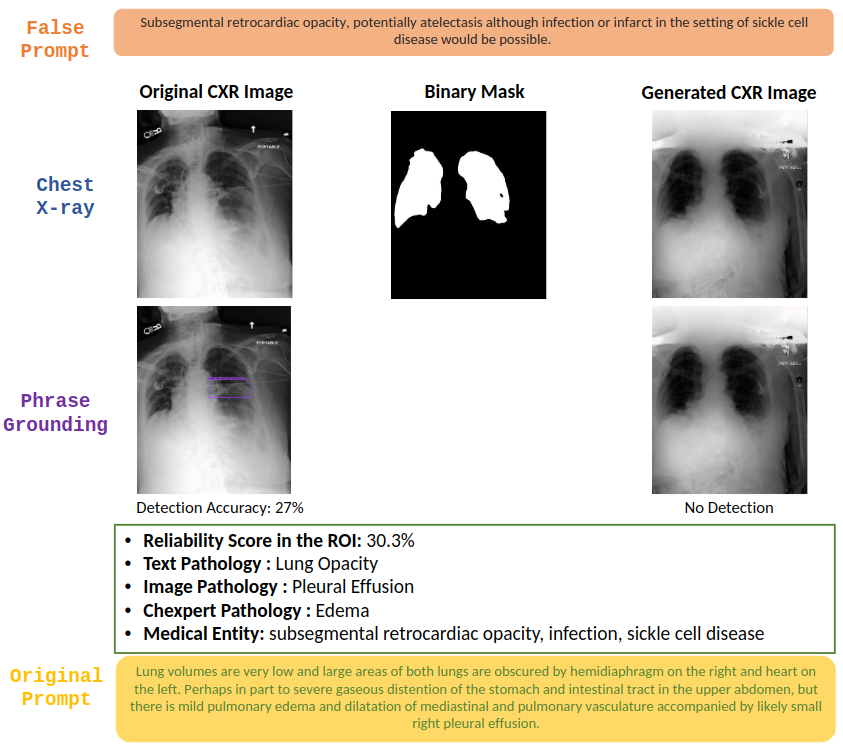}
    \caption{Case 4: A Success Case of Using False Report}
    \label{fig:case_4}
\end{figure}

In this case which is illustrated in Figure \ref{fig:case_4}, we examine a scenario where the prompt differs significantly from the original dataset prompt and shows no correlation with it. Although phrase grounding on the original CXR image identifies a region of interest (ROI) corresponding to the prompt, but no ROIs are detected when applying phrase grounding on the generated image.

Pathology detection from the text is entirely misaligned with the pathologies identified in the image, and the similarity score between the detected ROI in the original image and the corresponding region in the generated image is notably low. Furthermore, the annotated pathology from Chexpert differs substantially. For instance, while the annotation does not include ``pleural effusion" and instead notes only ``edema," the original report clearly indicates the presence of ``pleural effusion."

This case highlights two key insights: (1) even trusted annotations can occasionally lack accuracy or completeness, and (2) the model demonstrates robustness by not incorrectly aligning this mismatched prompt with the given CXR image. This underscores the importance of incorporating our VICCA approach to ensure reliability and trustworthiness in automated systems.

\subsection{Case 5: No Pathology}

\begin{figure}[t!]
    \centering
    \includegraphics[width=\linewidth]{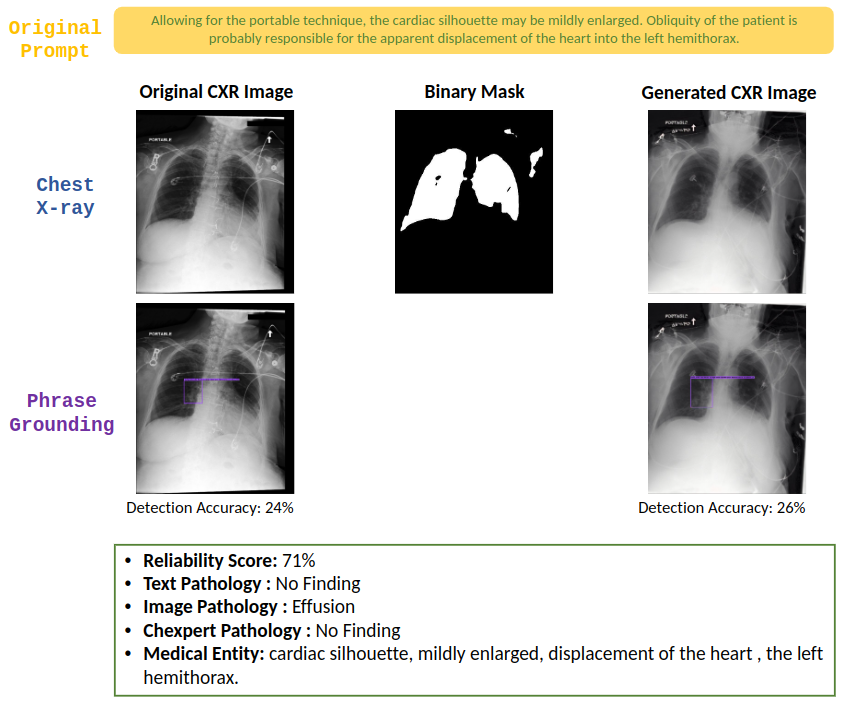}
    \caption{Case 5: No Pathology}
    \label{fig:case_5}    
\end{figure}

In our final case in Figure \ref{fig:case_5}, we examine an instance where the original annotation indicates no pathology. While the report describes an ROI in the image, our phrase grounding model detects an anomaly within the ROI in both the generated and original images. The reliability score indicates a high degree of similarity, but the detection accuracy score is notably low. Additionally, our image classification model identifies a pathology for this image.

Upon closer inspection of the text prompt, we observe a probable diagnosis linked to the ROI. Although the official annotation suggests no pathology, the findings from the VICCA model highlight a potential ambiguity. This case could be considered inconclusive, as the report itself is somewhat ambiguous, complicating the decision-making process.

\end{document}